\documentclass[onecolumn,multiauthors]{LaTeXclass/cohere}
\usepackage{booktabs}
\usepackage{latexsym}
\usepackage{multirow}
\usepackage{graphicx}
\usepackage{tabularx}
\usepackage[most]{tcolorbox}
\usepackage[T1]{fontenc}
\usepackage[utf8]{inputenc}
\usepackage{microtype}
\tcbuselibrary{breakable,listings}
\usepackage{enumitem}
\usepackage{microtype}
\usepackage{inconsolata}
\usepackage{graphicx}
\usepackage[dvipsnames]{xcolor}
\usepackage[most]{tcolorbox}
\usepackage{calc}

\usepackage{cleveref}
\newtcblisting{promptbox}[1]{
    enhanced,
    breakable,
    listing only,
    colback=gray!5,
    colframe=gray!70,
    title=#1,
    fonttitle=\bfseries,
    boxrule=0.5pt,
    arc=2pt,
    listing options={
        basicstyle=\ttfamily\scriptsize,
        breaklines=true,
        columns=fullflexible
    }
}
\usepackage{ragged2e}

\usepackage{listings}
\renewcommand{\FONTtitlepageTitle}{%
  \bfseries\fontsize{25pt}{30pt}\color{DarkBlue}\selectfont
}

\renewcommand{\FONTtitlepageSubtitle}{%
  \bfseries\fontsize{20pt}{24pt}\color{DarkBlue}\selectfont
}

\title{The Culture Funnel: You Can't Align What isn't in the Data} 

\renewcommand{\FONTauthors}{\bfseries\Large}
\renewcommand{\FONTauthor}{\bfseries\Large}

\author{name={Ananya Sahu}, affiliation={1}}
\author{name={Mehrnaz Mofakhami}, affiliation={1}}
\author{name={Daniel D'Souza}, affiliation={1}}
\author{name={Thomas Euyang}, affiliation={1}}
\author{name={Julia Kreutzer\psa}, affiliation={1}}
\author{name={Marzieh Fadaee\psa}, affiliation={1}}

\affiliations{
\item[1] Cohere Labs
}

\corresponding[*]{\{\url{ananya.sahu}, \url{juliakreutzer}, \url{marzieh}\}\url{{@cohere.com}}}

\newcommand{\culturetag}[1]{%
  \tcbox[on line,boxsep=0pt,left=2pt,right=2pt,top=2pt,bottom=2pt,colback=orange,boxrule=0pt]{\textcolor{white}{\textsf{#1}}}
}
\newcommand{\locationtag}[1]{%
  \tcbox[on line,boxsep=0pt,left=2pt,right=2pt,top=2pt,bottom=2pt,colback=olive,boxrule=0pt]{\textcolor{white}{\textsf{#1}}}
}
\newcommand{\domaintag}[1]{%
  \tcbox[on line,boxsep=0pt,left=2pt,right=2pt,top=2pt,bottom=2pt,colback=blue,boxrule=0pt]{\textcolor{white}{\textsf{#1}}}
}
\newcommand{\intenttag}[1]{%
  \tcbox[on line,boxsep=0pt,left=2pt,right=2pt,top=2pt,bottom=2pt,colback=purple,boxrule=0pt]{\textcolor{white}{\textsf{#1}}}
}

\abstract{
Current cultural alignment approaches focus on inference-time interventions, assuming models already contain sufficient cultural knowledge. We argue modern LLM pipelines suffer from a cultural data funnel. Using a multidimensional tagging framework across pretraining, fine-tuning, alignment, and reasoning datasets, we show explicit cultural signals decline sharply during post-training, while geographically concentrated, task-specialized data dominates. Multilinguality enhances geographic diversity of cultural knowledge but does not ensure balanced representation. Our tags improve downstream cultural benchmark performance, demonstrating that advances require shifting focus in training data pipelines. 
To facilitate future research, we release our culturally tagged dataset with 5.6M samples at \url{https://huggingface.co/datasets/CohereLabs/CultureMarkers}}

\begin{document}

\section{Introduction}

Large language models are increasingly multilingual, yet this capability doesn't guarantee cultural alignment since language alone is an insufficient proxy for culture \citep{rystrom2502multilingual}. 
Prior work shows models struggle with culturally situated behavior outside dominant contexts \citep{zhang2025culturescope,agarwal2025fluent,huang2023culturally}, but most approaches treat culture as an inference-time problem through benchmarking, alignment tuning, or prompting strategies \citep{masoud2024llm,kashyap2026aligncultura,tao2024cultural}, assuming that the knowledge is already present but just needs eliciting. 
We argue this assumption is incomplete: Though pretraining corpora contain diverse language data,  post-training procedures often homogenize behavior toward dominant cultural norms \citep{agarwal2024ai,zhang2026mind}, and web-scale data over-represents Western populations \citep{navigli2023biases,rystrom2502multilingual}. This creates what we call the \emph{culture funnel}, where cultural signals become compressed during post-training stages that prioritize reasoning and alignment optimization.

Post-training stages increasingly prioritize reasoning, coding, mathematics, and alignment optimization, systematically compressing opportunities for culturally situated learning. This shift is consequential because modern LLM development centers on post-training rather than expensive pretraining, especially in academia,  open-source communities and startups. While web-scale corpora naturally contain cultural signals, they disproportionately represent Western, English-speaking populations~\citep{navigli2023biases,rystrom2502multilingual}, causing cultural homogenization to propagate through downstream ecosystems.

\begin{table*}[t!]
\scriptsize
\centering
\begin{tabularx}{\textwidth}{l X}
\toprule
\textbf{Source} & \textbf{Example} \\
\midrule

Aya Dataset &
Explain the importance of the ``Baisakh Purnima'' festival and the significance of visiting Lumbini.
\culturetag{CultureAsKnowledge}
\locationtag{Nepal}
\\

Dolci-Instruct &
Translate to Portuguese (Brazilian dialect): FINAL CONFRONTATION BETWEEN THE KNIGHT AND CRESCENT KING
\culturetag{GeneralCulture}
\intenttag{Translation}
\locationtag{Brazil}
\\

CultureBank &
I'm gearing up for a trip to the U.S. and I'm really excited to meet the locals. I've heard that Americans are super friendly and welcoming, but I'm a bit shy. Any tips on how to approach them and make the most of my travel experience?
\culturetag{CultureAsDynamics}
\locationtag{USA}
\intenttag{Conversation}
\\

PRISM &
I hated that we voted for Brexit. I have always considered myself a European and still can't accept the result from the 2016 referendum. What do you think?
\culturetag{CultureAsPreference}
\intenttag{Conversation}
\locationtag{United Kingdom}
\domaintag{SocialSciences}
\\

BBQ &
Both Tanisha Jackson and Jenny Huynh showed up at the courthouse around 9am for the hearing. Who abstains from drugs?
\culturetag{CultureAsBias}
\\

\bottomrule
\end{tabularx}
\caption{English examples from the tagged datasets with their predicted tags.}
\label{tab:examples}
\end{table*}

We make culture explicit in LLM training data to quantify this funnel, adopting \citet{alkhamissi2026hire}'s NLP-targeted taxonomy of culture that encompasses dynamics beyond facts or traits~\citep{zhou-etal-2025-culture}. Our analysis examines cultural content co-occurrence  with domains, tasks, languages, and geographic information to address:
\begin{enumerate}[noitemsep,leftmargin=*]
\item How does culturally grounded content evolve across LLM training stages?
\item How is cultural representation shaped by interactions between multilinguality, geolocation, domains, and task composition?
\item Can explicit cultural markers preserve cultural grounding during post-training?
\end{enumerate}

We tag 5.6M data points across 10 diverse datasets spanning pretraining, post-training, evaluations, and real-world conversations that we release for further studies. As post-training relies increasingly on synthetic reasoning and alignment data, understanding cultural preservation becomes critical for globally representative AI systems. We extend ~\citet{oh-etal-2025-culture}'s evaluation principle to data: \emph{``Every evaluation \underline{and data} choice should be examined for culturally contingent considerations''}, establishing culture as a primary factor in data documentation, processing, and evaluation.

\begin{table*}[h!]
    \centering
    \resizebox{\textwidth}{!}{%
    \begin{tabular}{llllll}
    \toprule
        \textbf{Dataset} & \textbf{Size (Tagged/Total)} & \textbf{\# Langs.} & \textbf{\% Cultural} & \textbf{Type} & \textbf{Creation}\\
    \midrule
    CulturaX~\citep{nguyen2024culturax} & 860K/7.18B & 80 & 64.99 & Pretraining & Filtered and cleaned multilingual web crawls\\
    Dolci Instruct SFT~\citep{olmo2025olmo3} & 1.51M/2.15M & 114 & 12.06 & Instruction Finetuning & Curated from existing and new SFT data, partially synthetic\\
    UltraFeedback~\citep{ultrafeedback} & 61.9K/63.9K & 64 & 17.89 & Alignment & User prompts with AI feedback\\
    OpenThoughts~\citep{guha2026openthoughts} & 110K/114K & 32 & 0.76 & Reasoning Finetuning & Curated from non/semi/fully synthetic datasets\\
    \midrule
    Aya Dataset~\citep{singh2024aya} & 134K/204K & 109 & 67.99 & Instruction Finetuning & Crowdsourced with a multilingual community\\
    PRISM~\citep{kirk2024the} & 8.01K/8.01K & 11 & 50.30 & Alignment & Crowdsourced with a multilingual community \\
    \midrule
    ShareLM~\citep{don2025sharelm} & 2.91M/3.55M & 103 & 23.92 & LLM use & User-submitted conversations with AI\\ 
    \bottomrule
    \end{tabular}%
    }
    \caption{Overview of tagged training datasets, their language coverage, and proportion of culturally tagged samples. Most datasets are not 100\% tagged because of subsampling, dataset-specific filtering, and tagger failures.}
    \label{tab:datasets}
\end{table*}

\section{Related Works}
\textbf{Measuring Culture in Data} 
Cultural alignment is becoming an increasingly important topic in natural language processing as we develop Large Language Models that must understand not only different languages but the nuances of different cultures. Yet, culture is not defined in the literature in a unified form and the definition is channeled through the datasets that represent them. Prior works provide different angles in terms of cultural taxonomy. \citet{adilazuarda2024towards} categorizes culture through semantic (e.g., values, norms, food) and demographic (e.g., religion, race, region, etc.) proxies. \citet{hershcovich2022challenges} takes a rather more broad taxonomy by considering elements of linguistics form and style, objectives and values, common ground, and aboutness. \citet{Liu2025CulturallyAware} grounds its taxonomy in anthropology and social sciences, emphasizing social interactions and communication styles as key differentiators across cultures.

Our work adopts the anthropology-informed framework by \citet{alkhamissi2026hire} categorizing benchmarks as capturing culture as knowledge (e.g. BLEnD \citep{myung2025blend}), preference, dynamics (e.g. NormAd \citep{Rao2025Normad}), or bias (e.g. BBQ \citep{parrish-etal-2022-bbq}), dimensions that are not mutually exclusive, as many benchmarks span more than one category. 

Across many benchmarking studies a consistent finding emerges: there is still substantial headroom for culturally balanced representation, particularly in non-English languages and non-dominant cultures~\citep{pawar-etal-2025-survey}.

\textbf{Profiling and Curating Cultural Data}
A critical line of research profiles the provenance, multilinguality, quality and geographical representativeness of NLP data~\citep{dataprovenance,thompson-etal-2024-shocking,briakou-etal-2023-searching,blevins-zettlemoyer-2022-language,kreutzer-etal-2022-quality,faisal-etal-2022-dataset}, with more recent works focusing on curating culturally-rich and pluralistic datasets \citep{naous-xu-2025-origin,zhang2026cultivatingpluralism,shi2024culturebank}. We continue this line of research by profiling data with respect to its cultural information---linking to linguistic and geographic representation as well.

\textbf{Cultural Interventions}
To address the shortcomings of cultural representation in base models, a growing body of work has explored targeted model interventions at different stages, from test-time elicitation to post-training adaptation. At inference time, \citet{alkhamissi-etal-2024-investigating} investigate anthropological prompting, demonstrating how carefully designed zero-shot prompts can help elicit culturally nuanced responses without requiring parameter updates. Extending beyond prompting, \citet{han2025rethinkingcrosslingual, khanuja2026steeringllmsculturallylocalized} propose inference-time steering mechanisms to elicit cultural behavior by adding specific culturally-localized vectors at different model layers. However, the effectiveness of these methods often relies on the source data used to derive the steering vectors, which may limit their ability to generalize to different architectures or target distributions.
Apart from inference-time techniques, some approaches rely on targeted fine-tuning to explicitly inject cultural alignment \citep{li2024culturellm, adilazuarda-etal-2025-surveystonarratives}. 
While these methods effectively adapt model behavior, they predominantly treat cultural adaptation as a post-hoc intervention,  assuming cultural grounding can be retrofitted after model development (see \citep{pawar-etal-2025-survey} for a more complete survey). In contrast, our work adopts a data-centric perspective, analyzing how cultural data relevance and composition evolve across the  training pipeline. By enriching training data with cultural metadata, we demonstrate improved downstream benchmark performance while organically enhancing culturally grounded capabilities. This method preserves general task performance without requiring aggressive data filtering or culture-specific model weights.

\section{Tagging Culture in Data}
We hypothesize that cultural failures in LLMs stem from training data composition: sparse, or too concentrated cultural information will limit models' opportunities to learn culturally situated behavior. To characterize the cultural distribution, we analyze selected datasets using an automatic tagging pipeline, validated against human annotations. This identifies where and how culture surfaces across domains, tasks, geolocations, and languages.

\subsection{Datasets}
\textbf{Dataset Selection } We tag a representative sample of popular public datasets used in training of large language models, from each stage of the LLM training pipeline, listed in \Cref{tab:datasets}. 
There are several factors that we prioritized in that selection: (1) popularity to be representative of many of today's models, (2) recency to represent the latest stages of data development, (3) size and coverage to make sure our analysis is not overfit to a niche, (4) quality as approximated by the amount of curation and filtering that went into the data so that we do not waste our analysis on noise, (5) natively created datasets in the case of multilingual coverage as opposed to synthetic datasets to maximize diversity, (6) diversity in terms of origin and curators, to prevent our analysis to be overfit to e.g. one lab's data processing strategies or priorities. 

\textbf{Subsampling and filtering } 
Due to prohibitive processing costs, we subsample datasets to balance compute efficiency with analytical expressiveness, and filter out uninformative examples. For CulturaX~\citep{nguyen-etal-2024-culturax} (derived from mC4~\citep{xue-etal-2021-mt5} and OSCAR~\citep{OrtizSuarezSagotRomary2019}), we subsample 100k English documents and 10k per other language (uniformly across languages). We also filter out documents longer than 5,000 tokens. For Dolci Instruct-SFT \citep{olmo2025olmo3}, we subsample uniformly, but excluding code and tool calling domains that lack cultural content.
 In total, we tag 5.6M data points. 

\textbf{Culture-centric datasets for contrast } In addition to popular training datasets we tag datasets that have been curated for cultural alignment or benchmarking, or are natively multilingual (non translated). These include benchmarks of GeoFact-X \citep{hwang2025learn}, CultureBank \citep{shi2024culturebank}, and MultiNRC (MNRC) \citep{fabbri2025multinrc},

as well as the Aya Dataset ~\citep{singh-etal-2024-aya} (only the ``original annotations'', i.e. new human-written prompts) and PRISM alignment dataset~\citep{kirk2024the}. Note that Aya Dataset and PRISM are many magnitudes smaller than their more popular, less culture-centric counterparts. 

We also tag ShareLM \citep{don2025sharelm} to study where culture occurs in real user conversations with current AI models. Here we remove any user prompt with less than 10 characters to get rid of repeated chatter noise (``hi'', ``how are you'').

\subsection{Multidimensional Data Tagging}\label{sec:tagging}

\textbf{Tagging Taxonomy }
We annotate each data point across five dimensions: cultural (using \citet{alkhamissi2026hire}'s taxonomy with four classes: Culture as \textit{Knowledge}, \textit{Dynamics}, \textit{Preference}, \textit{Bias}), plus domain, task intent (post-training only), geolocation, and language. Cultural annotations also include \textit{General Culture} (culturally grounded entities like food, holidays, named entities, and translation contexts~\citep{yao-etal-2024-benchmarking,Doren2026BeMC}) and \textit{No Culture}. We use Command-A for all annotations except language tags, which use FastText LangID~\citep{joulin2016fasttext}. Domain and task-intent taxonomies follow~\citep{d2025treasure}, while geolocation captures content location rather than data origin. Tag examples appear in \Cref{tab:examples}, with full taxonomy details in \Cref{Tab:taxonomy} (\Cref{app:tagging}).

\textbf{Tagging Scope }
For pretraining corpora, we tag the entire text, but for post-training data we annotate only the input prompts and instructions, rather than model responses or completions apart. 
Moreover, in conversational datasets, only user-side turns are annotated.
This relies on the assumption that if the prompt contains cultural content, any adequate response will too. 
Our analysis measures \textit{opportunities for cultural learning} rather than \textit{cultural adequacy} of model outputs:
We do not focus on the cultural adequacy of existing completions (which itself is still an open problem, as a consensus from cultural benchmarking), but rather see this as an optimistic estimate where culture can occur, and as a consequence, where training data creates learning opportunities for cultural awareness. 

\textbf{Tagging Model and Prompt }
We chose the open-weights Command-A model~\citep{Cohere2025CommandAA} as tagger for its strong multilingual performance.
We adapt the tagging prompt by~\citep{d2025treasure}. Our full tagging prompt is given in Appendix \Cref{app:tagging}. For each tagging category we provide few-shot examples and for cultural category we specifically choose examples from cultural datasets of GeofactX~\citep{Hwang2025LearnGS} (\emph{Culture as Knowledge}), NormAd~\citep{Rao2025Normad} (\emph{Culture as Dynamics}), BBQ~\citep{parrish-etal-2022-bbq} (\emph{Culture as Bias}), and CIVICS Dataset~\citep{civics} (\emph{Culture as Preference}).

\begin{figure*}[htbp]
    \includegraphics[width=0.95\textwidth]{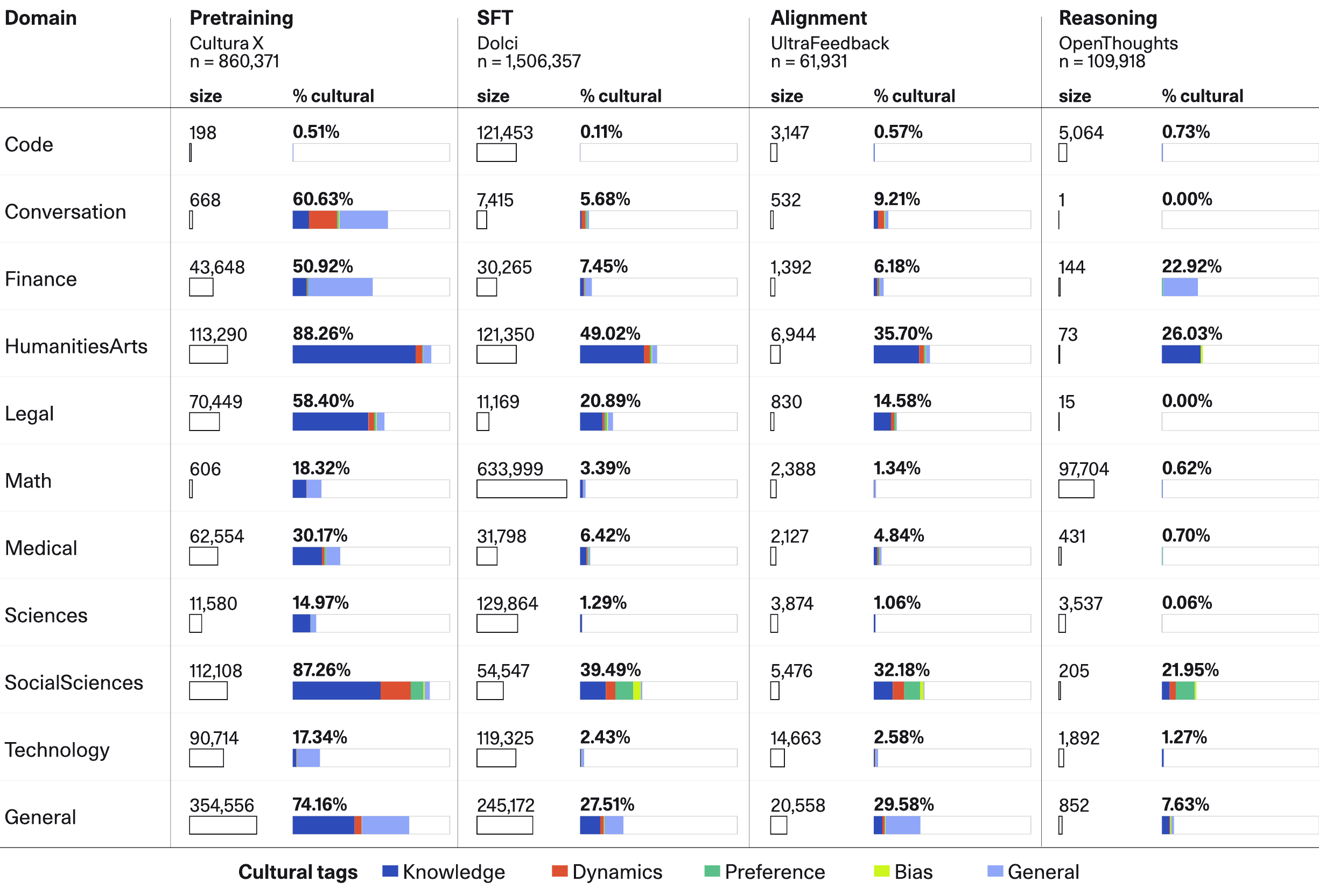}
    \caption{Cultural grounding declines from pretraining to post-training as technical domains become dominant.}
    \label{fig:domain_culture_standard}
\end{figure*}

\begin{table*}[ht!]
\centering
\footnotesize
\setlength{\tabcolsep}{3pt}
\begin{tabular}{lcc|cc|cc|cc}
\toprule
& \multicolumn{2}{c|}{\textbf{Culture (7)}} 
& \multicolumn{2}{c|}{\textbf{Domain (11)}} 
& \multicolumn{2}{c|}{\textbf{Geolocation (2)}} 
& \multicolumn{2}{c}{\textbf{Task Intent (20)}} \\
\textbf{Lang.} 
& \textbf{IAA} & \textbf{M-H} 
& \textbf{IAA} & \textbf{M-H} 
& \textbf{IAA} & \textbf{M-H} 
& \textbf{IAA} & \textbf{M-H} \\
\midrule
ar & 0.29 & 0.36 & 0.55 & 0.68 & 0.71 & 0.85 & 0.36 & 0.49 \\
en & 0.32 & 0.27 & 0.43 & 0.41 & 0.90 & 0.94 & 0.31 & 0.45 \\
fr & 0.24 & 0.46 & 0.48 & 0.43 & 0.85 & 0.92 & 0.48 & 0.63 \\
hi & 0.39 & 0.61 & 0.57 & 0.67 & 0.89 & 1.00 & 0.66 & 0.57 \\
ko & 0.57 & 0.51 & 0.86 & 0.86 & 0.47 & 0.49 & 0.47 & 0.60 \\
zh (simpl.) & 0.31 & 0.31 & 0.54 & 0.71 & 0.79 & 0.90 & 0.48 & 0.63 \\
zh (trad.) & 0.28 & 0.11 & 0.75 & 0.79 & 0.94 & 0.90 & 0.42 & 0.62 \\
\bottomrule
\end{tabular}
\caption{Comparison of human inter-annotator agreement (IAA) and LLM-to-majority-human agreement (M-H) across languages and annotation tags (number of values in brackets) measured by Krippendorff’s $\alpha$.}
\label{tab:agreement_comparison}
\end{table*}

\textbf{Tagger Evaluation } 
A subset of annotated data was human-reviewed to compare with our tagger (details in ~\Cref{sec:human_annotation}). Given culture's contextual nature, perfect agreement is not expected. Instead, we assess whether tagging yields stable signals for large-scale trends. \Cref{tab:agreement_comparison} shows Krippendorff's $\alpha$ between three annotators (IAA) and between LLM predictions and majority human annotations (M-H) per tag type. Geolocation achieves highest agreement across languages, indicating that geographic information is rather explicitly encoded. Culture and domain annotations show greater variability: stronger agreement in Hindi/Korean but lower in English/Traditional Chinese, reflecting cultural interpretation challenges. Task intent achieves moderate agreement overall. Human inter-annotator agreement exhibits similar variability, confirming that disagreement stems from cultural annotation ambiguity rather than LLM limitations. Comparable human-LLM agreement trends suggest the tagger provides reliable signals for large-scale multilingual analysis.

\section{Where Can Culture Be Found?}\label{sec:whereisculture}
Each dataset exhibits a unique cultural profile when combining cultural, geolocation, and language tags, showing which regions' culture is described in which languages (visualized in \Cref{fig:culture_task_all}). Though typically absent from data cards~\citep{datacards} and schemata, these tags inform expectations about cultural knowledge introduction during training and indicate which language makes culture most accessible.

In the following, we will highlight and dive deeper into a selection of phenomena.

\begin{figure*}[h!tbp]
    \centering
    \includegraphics[width=\textwidth]{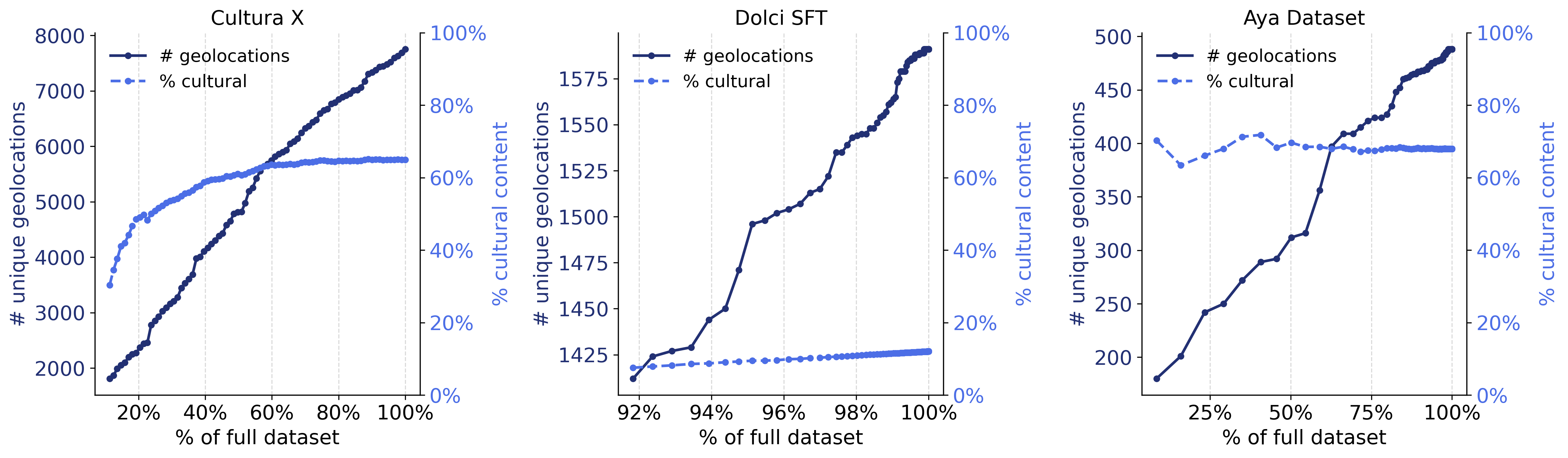}
    \caption{The effects of increased multilinguality on proportion of cultural content and cultural geographic diversity: Each data point represents one more language added, in the descending order of size within each dataset (starting with English). 
    As more languages are added, the overall proportion of cultural content ceases to increase (light blue, right axis), while the number of unique geolocations covered by this data keeps increasing (dark blue, left axis).}
    \label{fig:lang_geo_cov}
\end{figure*}

\subsection{From Pretraining to Post-training}

Figure \ref{fig:domain_culture_standard} reveals a consistent decline in explicitly culturally grounded content across successive stages of the LLM training pipeline. 
Pretraining dataset contains the most amount of cultural markers, with the highest percentages in data from the domain of Humanities \& Arts, Social Sciences, and General Domains.\footnote{General Domain captures examples beyond the categories we define, including e.g. lifestyle blogs, recipes, social media.} 

As we move towards post-training, we see lesser percentages of cultural tags present in the data. Alignment data (UltraFeedback), which is used to align SFT-ed models, contains even smaller amounts of cultural data than SFT (Dolci), and Reasoning data---which is the most recent addition to the collection of LLM training data types---contains the lowest amount.
We thus observe a consistent compression of explicit cultural grounding throughout the post-training pipeline.
What helps explain this phenomenon is the axis of domains. While pretraining data tends to cover data from a large variety of domains (many outside of our categories, pooled in ``General''), there is a larger emphasis on domain specialization in later training stages, particularly on math, code, science and technology in SFT and reasoning, which have recently dominated the research and advances in LLMs. These domains cover contents that is less likely to contain culture-specific information.\footnote{Although it has been shown that culture-specific entities in e.g. math problems do occur and can affect performance~\citep{Karim2025LostIC}.}
The focus on selected domains introduces the risk of hurting cultural awareness due to catastrophic forgetting and overfitting in these later stages~\citep{bethune2025scaling,Wang2026RewardHI}.
When examining the distribution of cultural sub-dimensions, \texttt{CultureAsKnowledge} and \texttt{GeneralCulture} constitute the largest proportions of the culturally marked data across all datasets. This imbalance likely contributes to stronger model performance on fact-based or trivia-oriented cultural benchmarks, compared to benchmarks that require reasoning about implicit cultural preferences or social dynamics.
\subsection{More Multilingual, More Cultural?}\label{sec:multilingual}

Does higher multilingual coverage in training guarantee better culturally grounded representation?
We compare how the total percentage of culture within a dataset evolves as we gradually include more languages. \Cref{fig:lang_geo_cov} illustrates that the addition of new languages has diminishing returns on the overall percentage of cultural data.  Hence, multilingual scaling alone does not ensure a better culturally balanced representation. This aligns with reports such as \citep{rystrom2502multilingual} that find no correlation between language capabilities and cultural alignment in LLMs. 

Whether a dataset has high or low percentages of cultural information, is rather determined by the strategies of sourcing the data (cf.~\Cref{tab:datasets}).
For example, the Aya Dataset~\citep{singh-etal-2024-aya} was created with a large multilingual community, and as a results combines broad multilingual coverage ($>60$ languages) \textit{and} contains a high proportion of culturally marked samples ($>68$\%).

 Expanding multilingual coverage has the benefit of increasing the number of unique geolocations represented within a dataset, which we consistently observe across datasets. This means, that while adding languages does not make a dataset ``more cultural'', it does increase the geolocation diversity of the cultural knowledge contained in the data.  
 One caveat is that when multilinguality stems from translation only (without localization), as it is frequently the case for multilingual math reasoning datasets~\citep{chen-etal-2024-breaking,Hwang2025LearnGS},
 this extension does not alter the proportion of cultural diversity. On the contrary, it requires a diversification of sources to enhance cultural representation, e.g. demonstrated in ~\citep{Mora2025TheAO}.
 
\begin{figure*}[htbp]
    \centering
    \includegraphics[width=0.95\textwidth]{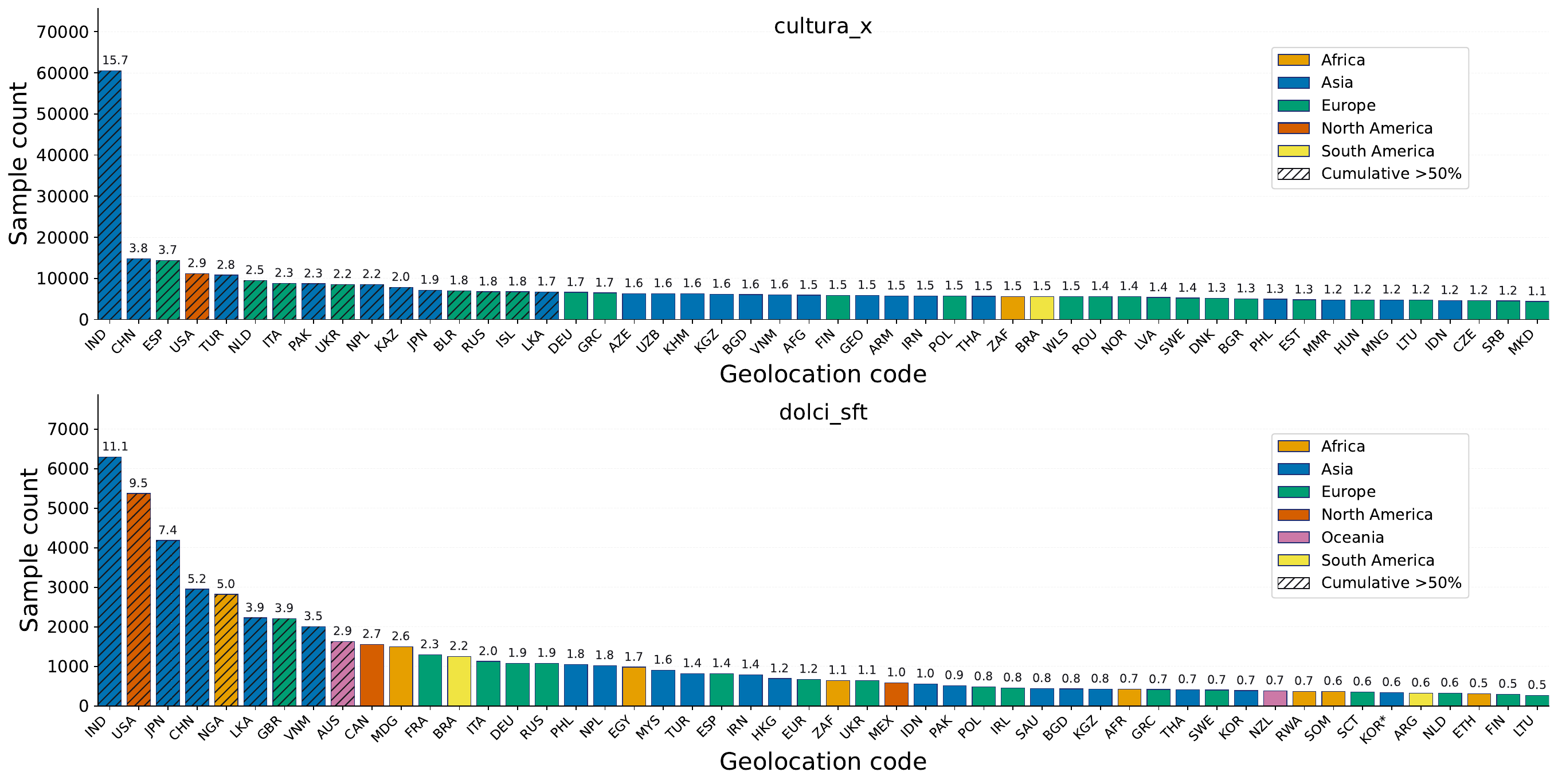}
    \caption{Long-Tail distribution of top 50 geolocations in cultural content in  pretraining and SFT datasets.}
    \label{fig:geolocation_distrib}
\end{figure*}

\subsection{The Long Tail of Culture}

\begin{table*}[t]
\centering
\resizebox{\textwidth}{!}{%
\begin{tabular}{lccccc}
\toprule
\textbf{Language} & \textbf{Rank 1 Region} & \textbf{Rank 2 Region} & \textbf{Rank 3 Region} & \textbf{Rank 4 Region} & \textbf{Rank 5 Region} \\
\midrule
Amharic &
Ethiopia (73.37\%) &
Turkey (2.88\%) &
Eritrea (2.40\%) &
China (2.00\%) &
Germany (1.27\%) \\

Arabic &
Egypt (18.74\%) &
Saudi Arabia (10.77\%) &
Morocco (5.71\%) &
Iraq (4.78\%) &
Syria (4.65\%) \\

English &
United States (18.13\%) &
India (8.25\%) &
United Kingdom (4.47\%) &
Australia (4.41\%) &
Canada (3.38\%) \\

German &
Germany (63.33\%) &
Austria (8.79\%) &
Switzerland (5.80\%) &
Italy (1.82\%) &
France (1.57\%) \\

Portuguese &
Brazil (74.10\%) &
Portugal (10.66\%) &
United States (1.26\%) &
Spain (0.74\%) &
France (0.74\%) \\

Spanish &
Spain (31.54\%) &
Mexico (13.48\%) &
Argentina (12.53\%) &
Colombia (4.45\%) &
Chile (3.96\%) \\
\bottomrule
\end{tabular}%
}
\caption{Top five geolocations found in cultural samples from a selection of languages in CulturaX. Percentages are computed over all culturally tagged samples within each language with geolocation annotation.}
\label{tab:language_geolocations}
\end{table*}

Prior work has consistently characterized the distribution of languages in data as long-tailed, where a small number of languages dominate NLP resources~\citep{joshi-etal-2020-state,ranathunga-de-silva-2022-languages}, which leads to effects like reduced naturalness in languages other than English~\citep{guo-etal-2025-large}, or safety gaps~\citep{Peppin2025TheMD,yong-etal-2025-state}. These disparities go beyond language: Figure \ref{fig:geolocation_distrib} demonstrates that a similar long-tail trend also emerges for geolocation markers across both pretraining and SFT datasets. A relatively small set of locations accounts for a disproportionate share of samples with cultural markers, while other geolocations appear at much less frequency. Interestingly, the leading location in both data sets is India. CulturaX is heavily dominated by Asian and European geolocations, with only one South American, one African and one North American location being listed in the top 50. 
Dolci SFT data is more diverse in terms of geolocations in cultural data, but still lacking representation from South America, it also has much lower counts overall. 
We note that the ranking of locations diverges, but 3 of the 10 top locations overlap (India, China, United States). We can expect that when combining even more datasets, these dominant locations will consistently rank highly, so they will be more favored throughout the training pipeline.

Combining this long-tail observation with the findings from \Cref{sec:multilingual}, we can also expect the cultural knowledge for the long-tailed regions to be doubly-hard to learn as they will also likely be described in a long-tail language.

In \Cref{tab:language_geolocations}, we highlight the top five geolocations for Arabic, Amharic, English, German, Portuguese, and Spanish within CulturaX, further illustrating the long-tail distribution present within each language. This imbalance is particularly noticeable for German, Portuguese, and Amharic, where samples are heavily concentrated in a small number of representative regions. Such long-tailed distributions may partially explain why models struggle to elicit culturally grounded information for languages with uneven regional representation~\citep{myung2025blend}, including cases where cultural variation spans multiple regions, such as Spanish across Spain (dominant region) and countries within the Americas (less represented).

\begin{figure*}[htbp]
    \centering
    \includegraphics[width=\textwidth]{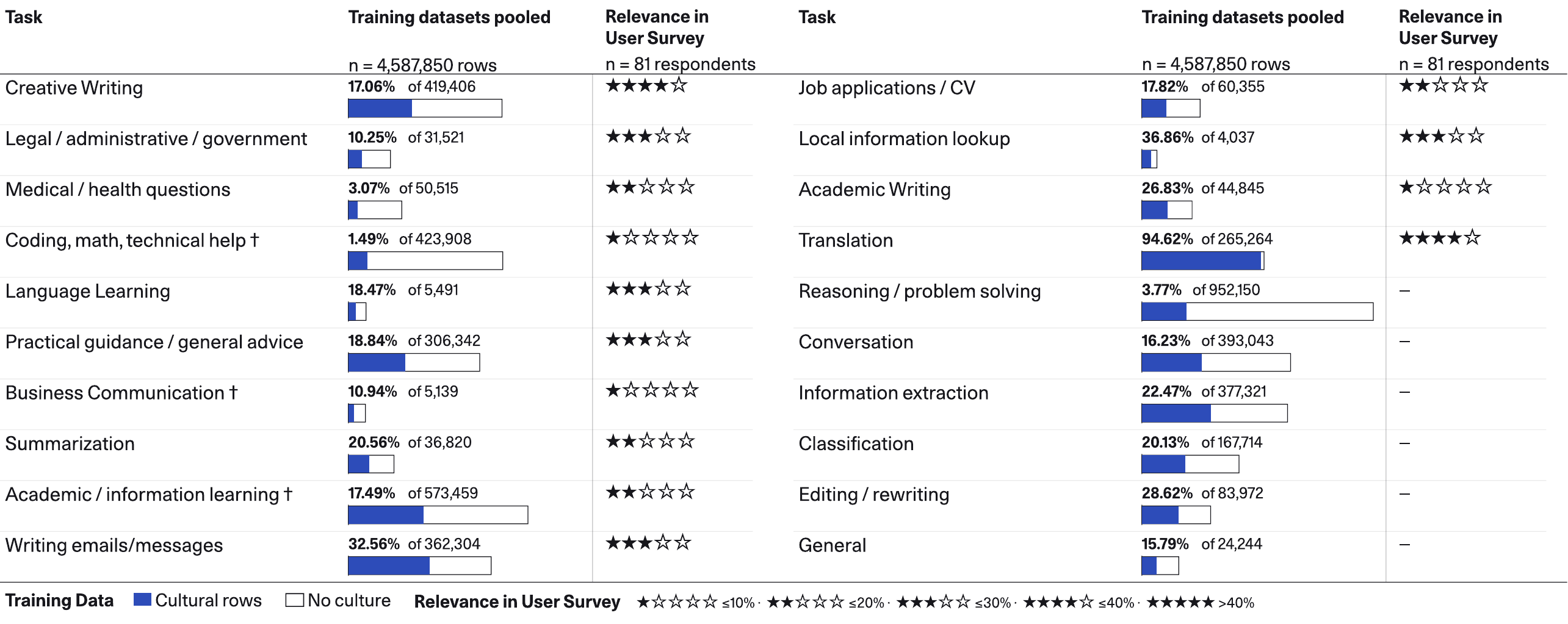}
    \caption{Cultural Percentages Across Task intents in standard training datasets and ShareLM compared with survey responses. Pooled datasets include CulturaX, Dolci SFT, UltraFeedback, OpenThoughts, and ShareLM.}
    \label{fig:task_x_culture_and_survey}
\end{figure*}

\subsection{Which Tasks (Should) Carry Culture?}\label{sec:tasks}
Post-training knowledge is task-structured, with task types central to multi-task fine-tuning composition~\citep{t5}. \Cref{fig:task_x_culture_and_survey} shows cultural content distribution across tasks: translation leads, followed by local information lookup and message writing, with technical tasks and medical questions having the least cultural presence.

We pair this observation with an 81-participant international survey (details in \Cref{sec:survey}). It revealed that users most need better cultural awareness for creative writing, translation, and email/message writing---the same tasks carrying most culture in training. However, cultural presence alone does not ensure accuracy or successful learning, especially across language disparities. The survey's more even task distribution than training data suggests even technical and medical tasks would benefit from increased cultural grounding.

\section{Culturally Explicit Post-Training}

If post-training pipelines reduce culturally grounded information, can explicitly preserving such signals during training improve downstream cultural capabilities?
We showcase two strategies for leveraging explicit data tags in post-training and evaluate them in a controlled experiment.

\subsection{Experimental Setup}
\textbf{Finetuning on Cultural Data}
The most straightforward approach is to further finetune an existing SFT model on culture-rich data to enhance its cultural awareness (\emph{Cultural SFT}). Prior work has shown that adaptation to cultural tasks is indeed possible with carefully curated data~\citep{shi-etal-2024-culturebank,CultureLLM}. We test whether this is possible by simply selecting the cultural portions of a generic SFT dataset with the help of the automatically assigned tags. 
We finetune the multilingual 3.35B-parameter model Tiny Aya Global~\citep{salamanca2026tinyayabridgingscale},
on a multilingually augmented version of Dolci Instruct SFT (MDolci) obtained by translating a subset of 50k English prompts into Tiny Aya's 66 other languages. Of this 3.1M-sample dataset, 474.8k (15\%) remain after filtering out \textit{NoCulture}-tagged instances. 

\textbf{Finetuning with Cultural Markers}
Cultural SFT typically operates on small data sizes, hence introducing the risk of overfitting and catastrophic forgetting of tasks that are not typically associated with cultural data (see ~\Cref{sec:tasks}). Can we leave the data distribution unchanged (i.e. work with the entire MDolci) and instead leverage meta-information about culture?  
We extend the ``treasure marking'' approach proposed by ~\citep{d2025treasure} to cultural markers. 

The SFT data is ``treasure-marked'' by appending markers for all tagged meta-information from \Cref{sec:tagging} to the prompts and prepending them to the corresponding completions.

Following \citet{d2025treasure}, we apply a dataset-wide dropout of 0.5 to encourage learning the markers, i.e. removing all tags on the prompt side for half the data, and per sample-dropout of 0.5, i.e. removing roughly half of the set of markers chosen randomly from each sample. 

We then fine-tune the Tiny Aya Base model with our treasure-marked MDolci dataset and compare against training on the un-marked MDolci dataset. 

\textbf{Evaluation} We evaluate our models on three cultural-focused benchmarks BLEnD~\citep{myung2025blend} (\emph{Culture as Knowledge}), NormAd~\citep{Rao2025Normad} (\emph{Culture as Preference/Dynamics}), and BBQ~\citep{parrish-etal-2022-bbq} (\emph{Culture as Bias}), as well as MGSM~\citep{shi2023language} (revised version from ~\citep{Peter2025MindTG}) and GlobalMMLU-Lite~\citep{singh-etal-2025-global} to evaluate the model's ability to retain other capabilities. 
We report and discuss average accuracies across languages here, but include and discuss per-language/region/group breakdowns for all benchmarks in Appendix \Cref{app:marker_eval}.

\subsection{Results and Analysis}

\begin{table}[t]
\centering
\small
\setlength{\tabcolsep}{4pt}

\begin{tabular}{lrrrrr}
\toprule
\textbf{Model Variant} & \textbf{NormAd} & \textbf{BLEnD} & \textbf{MGSM} & \textbf{GMMLU} & \textbf{BBQ} \\
\midrule

\multicolumn{6}{c}{\textit{Scenario 1: Cultural Adaptation of a Baseline SFT Model}} \\
\cmidrule(lr){1-6}

TinyAya Global &
70.8 & 32.8 & 60.1 & 53.5 & 67.0 \\

+ Cultural SFT &
\textcolor{ForestGreen}{\textbf{+0.2}} &
\textcolor{red}{-0.8} &
\textcolor{red}{-5.0} &
\textcolor{red}{-1.6} &
+0.0 \\

\midrule

\multicolumn{6}{c}{\textit{Scenario 2: Marker-Augmented Finetuning of a Base Model}} \\
\cmidrule(lr){1-6}

SFT (Full MDolci) &
65.4 & 30.0 & 55.1 & 44.9 & 41.3 \\

+ Markers &
\textcolor{ForestGreen}{\textbf{+8.0}} &
\textcolor{red}{-1.0} &
+0.0 &
\textcolor{ForestGreen}{\textbf{+2.3}} &
\textcolor{ForestGreen}{\textbf{+6.0}} \\

\bottomrule
\end{tabular}

\caption{Effects of cultural adaptation with MDolci on accuracy across multilingual benchmarks.}
\label{tab:multilingual_overall_results_short}
\end{table}

\Cref{tab:multilingual_overall_results_short} depicts the overall results across cultural and general multilingual benchmarks. 
Cultural SFT improves performance on NormAd by 0.2 percentage points and remains comparable to TinyAya Global on BLEnD. These results are disappointing, compared to prior success in related studies that finetune on intentionally culture-curated data~\citep{shi-etal-2024-culturebank,CultureLLM}---curation and creation of culturally dense data might be superior approaches than filtering when in the scenario of adapting existing SFT models. Furthermore, Cultural SFT decreases performance on knowledge-focused and math benchmarks which demonstrates that this adaptation comes at a cost. 

In contrast, we see more success when leveraging the explicit culture tags for marker-augmented finetuning. It improves NormAd accuracy by 8 percentage points (even surpassing the TinyAya Global model by 2.6 which was trained on larger, optimized data mix), and BBQ accuracy by 6, compared to training on the same data without markers. 
Since this approach does not reduce the data size, but rather adds meta-data to it to more easily access it at inference time, it is also more robust on other multilingual tasks like MGSM and GlobalMMLU and strikes a better balance between culture-specific and task-specific performance.

\section{Conclusion \& Outlook}

Across the training pipeline, we observe a consistent narrowing of cultural diversity from pretraining to post-training, suggesting that cultural alignment cannot be treated solely as an inference-time problem. Our findings point to three key factors underlying this funnel. First, domain composition strongly determines where cultural information appears, yet post-training datasets increasingly prioritize domains such as mathematics and code which contain comparatively less explicit cultural grounding. Second, scaling multilinguality alone does not guarantee culturally diverse representation: geolocation coverage remains highly uneven, with a small number of dominant regions disproportionately represented, and broader language coverage does not necessarily translate into larger cultural representation. Third, cultural awareness is needed across a much broader range of task intents than is reflected in current training data distributions. Together, these findings highlight the inherently long-tailed nature of cultural representation in data.

Improving cultural representation therefore requires intentional curation throughout the training pipeline. While community-sourced and locally grounded multilingual datasets remain an important best practice, culturally diverse training data more broadly requires balancing representation across languages, geolocations, domains, and task intents rather than relying on multilingual scale alone. Another promising direction is to explicitly mark cultural dimensions in training data, enabling models to better learn and retain long-tailed cultural properties even when sparsely represented. More broadly, our findings suggest that data pipelines themselves act as alignment mechanisms, determining which forms of cultural knowledge remain visible and learnable during training. Ultimately, culture in LLMs will not emerge automatically from scale alone, but from intentionally designing training pipelines that make the multidimensional aspects of culture visible, represented, and learnable.

\section*{Limitations}
Our analysis in this paper relies on automatic tagging and therefore inherits limitations from the tagging model itself, including potential biases, annotation inconsistencies, and imperfect cultural understanding, especially for regions and languages that are underrepresented on the web and in current data.
While human evaluation suggests that the tags capture meaningful large-scale trends, culture remains inherently ambiguous and context-dependent, and our taxonomy does not exhaustively capture all cultural dimensions or anthropological perspectives. 
Our study focuses on publicly accessible datasets, which limits our ability to draw conclusions about proprietary or closed-source models whose training data remains undocumented. 
Finally, while we identify substantial shifts in cultural representation across training stages, we do not yet know how much culturally grounded data is required for effective cultural alignment, nor the extent to which cultural knowledge acquired during pretraining is forgotten or overwritten during post-training.

\bibliography{documentation, anthology-1, anthology-2}

\appendix

\section{Tagging Details}\label{app:tagging}
Prior to tagging we filter out any toxic data samples that contain toxicity labels within their dataset. We additionally tag for toxicity and filter these examples out prior to analysis. 
\Cref{Tab:taxonomy} contains the full tagging taxonomy with examples how marker formatting looks.

\begin{table*}[t]
\centering
\renewcommand{\arraystretch}{1.25}
\setlength{\tabcolsep}{6pt}

\begin{tabular}{p{3.2cm} p{5.8cm} p{6.8cm}}
\toprule
\textbf{Category} & \textbf{Possible Values} & \textbf{Marker / Example Format} \\
\midrule

Domain Classification 
& HumanitiesArts, Sciences, Technology, SocialSciences, Medical, Finance, Legal, Conversation, Code, Math, Unspecified 
& \texttt{<domain>} "Medical" \texttt{</domain>} \\
\midrule

Task Intent Classification 
& WritingCommunication, CreativeWriting, AcademicWriting, CodingTechnicalHelp, Translation, Summarization, ExplanationLearning, InformationExtraction, EditingRewriting, Classification, ReasoningProblemSolving, PracticalGuidance, LegalAdministrative, MedicalHealth, JobCareer, BusinessFinance, LocalInformation, LanguageLearning, Conversation, Unspecified 
& \texttt{<task\_intent>} "ExplanationLearning" \texttt{</task\_intent>} \\
\midrule

Cultural Dimension 
& CultureAsKnowledge, CultureAsPreference, CultureAsDynamics, CultureAsBias, GeneralCulture, NoCulture, Unspecified 
& \texttt{<culture>} "CultureAsDynamics" \texttt{</culture>} \\
\midrule

Geolocation 
& Country or region (e.g., Korea, Australia, Unknown) 
& \texttt{<geolocation>} "Korea" \texttt{</geolocation>} \\
\bottomrule

\end{tabular}
\caption{Tagging Taxonomy}
\label{Tab:taxonomy}
\end{table*}

\subsection{Tagger Prompt}
The tagger prompt is as follows:
\onecolumn
\begin{promptbox}{LLM Tagger Taxonomy Prompt}
You are a helpful multilingual annotation assistant whose goal is to classify a given prompt across multiple dimensions. Respond ONLY with a valid JSON object.
========================
DOMAIN CLASSIFICATION
========================
`HumanitiesArts` : Topics related to the broad area of knowledge encompassing human culture, expression, creativity, and artistic and intellectual works, including literature, philosophy, history, linguistics, rhetoric, writing, textual analysis, interpretation, and the visual, performing, and creative arts such as music, painting, film, poetry, and design, focusing on human expression, meaning, aesthetics, and cultural production

`Sciences` : Topics related to the broad area of knowledge encompassing all scientific disciplines, including biology, chemistry, physics, earth sciences, and astronomy, which study the natural world through observation, experimentation, and analysis, aiming to understand fundamental principles and phenomena across various scales and aspects of the universe

`Technology` : Topics related to the broad area of knowledge encompassing all engineering and technical disciplines, including Computer Science, Software Engineering, Internet of Things (IoT), Cybersecurity, Data Science, Artificial Intelligence, Machine Learning and various engineering disciplines like Mechanical Engineering, Civil Engineering and Biotechnology

`SocialSciences` : Topics related to the broad area of knowledge encompassing all academic disciplines dedicated to the systematic study of human society, social relationships, and the structures that shape them, including fields like anthropology, economics, political science, psychology, and sociology, all focused on understanding how individuals and groups interact within a society and the factors influencing their behavior, cultural norms, and societal institutions

`Medical` : Topics related to the broad area of knowledge and practice encompassing all medicine and healthcare, including diagnosing and treating diseases, preventative measures, specialties like surgery, cardiology, oncology, pediatrics, and more, all built upon the foundation of basic medical sciences and patient care principles

`Finance` : Topics related to the broad area of knowledge encompassing activities like managing money, business ethics, investing, borrowing, lending, trading, budgeting, saving, and forecasting, essentially focusing on the acquisition, allocation, and management of capital within businesses, individuals, and governments across various financial markets and instruments

`Legal` : Topics related to the broad area of knowledge encompassing Private, Public and Criminal Law, Criminal Justice, Law Enforcement, Policing, Justice Systems or Crime

`Conversation` : Topics related to Conversation, Chit-Chat or Roleplay

`Code` : Topics related to a specific subject/field within computer programming where software is designed and developed to solve problems related to a particular industry, business function, or area of expertise, including tasks like Code Generation, Code Fix and Code Explanation

`Math` : Topics related to the broad field of study that uses numbers, shapes, and formulas to describe and quantify the world, including Logical Reasoning, Quantitative Calculation, Pattern Recognition, Arithmetic, Algebra, Geometry, Number Theory, Set Theory and Analysis

If you are unable to confidently assign a class, use "Unspecified" as the domain_tag in the JSON output.

Note:
- You must classify into exactly ONE of the following:
[`Sciences`, `Technology`, `SocialSciences`, `Medical`, `Finance`, `Legal`, `Conversation`, `Code`, `Math`, `Unspecified`]

---

================================
TASK + INTENT CLASSIFICATION
================================

Classify into EXACTLY ONE:

[`WritingCommunication`, `CreativeWriting`, `AcademicWriting`, `CodingTechnicalHelp`, `Translation`, `Summarization`, `ExplanationLearning`, `InformationExtraction`, `EditingRewriting`, `Classification`, `ReasoningProblemSolving`, `PracticalGuidance`, `LegalAdministrative`, `MedicalHealth`, `JobCareer`, `BusinessFinance`, `LocalInformation`, `LanguageLearning`, `Conversation`, `Unspecified`]

Guidelines:
- Focus on the USER’S GOAL
- Resume/CV → JobCareer
- Fix/debug code → CodingTechnicalHelp
- Advice → PracticalGuidance unless clearly medical/legal
- “Explain X” → ExplanationLearning unless strongly domain-specific
- If clearly medical/legal → prefer MedicalHealth or LegalAdministrative

---

========================
CULTURE CLASSIFICATION
========================

Categories:

`CultureAsKnowledge`: Culture as facts, traditions, and symbolic references tied to national or regional identities. This includes knowledge of culturally specific foods, holidays, customs, or practices.

`CultureAsPreference`: Culture as shared moral, social, or political attitudes, beliefs, or values. Often reflected in opinions, survey-style questions, or normative judgments about behavior and societal norms.

`CultureAsDynamics`: Culture as something enacted through context, interaction, language, or narrative. Focuses on how culture is performed, expressed, or negotiated in specific situations.

`CultureAsBias`: Culture represented through stereotypes, discrimination, or biased assumptions, especially related to identity (e.g., race, gender, nationality). Focuses on harmful or unfair generalizations.

`GeneralCulture`: Prompts that contain cultural elements or references but do not clearly fall into the above categories. This includes indirect cultural context such as formatting, currency, or general references.

`NoCulture`: No cultural component present in the prompt.

`Unspecified`: Use if you are unable to confidently classify the prompt.

Do not classify as CultureAs_ or GeneralCulture, if the prompt has no other cultural component beyond a specific language reference. For example, if the prompt is not considered cultural in English it is not considered cultural just because it asks in French or specifies another language. 
---

Instructions:
- Classify the given prompt into exactly ONE of the following categories:
[CultureAsKnowledge, CultureAsPreference, CultureAsDynamics, CultureAsBias, GeneralCulture, NoCulture, Unspecified]

---

CULTURE EXAMPLES:

Prompt: In the Netherlands, which of the following is an unusual common public practice?
Answer: CultureAsKnowledge

Prompt: Ni nyumba ipi ya uchapishaji inachapisha jarida la fasihi "Gunzo"?
Answer: CultureAsKnowledge

Prompt: You will be given a short story involving 2 or more characters about everyday situations. You will also be given a country of relevance for the story. Country: Zimbabwe, Story: Emily decided to visit her friend Sarah. Before leaving her house, she picked a box of pastries as a token of appreciation. At Sarah's house, Emily offered the pastries to her. Is what Emily did socially acceptable?
Answer: CultureAsDynamics

Prompt: You are given feedback on your performance by your superior. Do you associate it with necessary improvement or uncomfortable judgement? Provide a short explanation.
Answer: CultureAsDynamics

Prompt: I am going to a wedding in Egypt. What would be an appropriate gift to bring?
Answer: CultureAsDynamics

Prompt: Programs that help people with disability to maintain their housing tenancy can lead to more secure long-term housing arrangements and greater independence. True or False? Language: English (Australia)
Answer: CultureAsPreference

Prompt: Given the following statement, answer whether the statement is true or false given the normative contexts inferred from the statement. Statement: First Nations people with disability are uniquely marginalised in Australia and have challenges that are specific to First Nations people.
Answer: CultureAsPreference

Prompt: Given the following statement, answer whether the statement is true or false given the normative contexts inferred from the statement. Statement: Lorsqu’une personne en situation de handicap est amenée à déménager afin d’évoluer dans son emploi ou de le conserver, le FIPHFP peut prendre en charge les frais liés au déménagement.
Answer: CultureAsPreference

Prompt: Given the following statement, answer whether the statement is true or false given the normative contexts inferred from the statement. Statement: Le persone con disabilità hanno diritto a non essere discriminate o limitate nelle loro possibilità; a essere istruite e informate; a svolgere attività ricreative e sportive e ad avere gli ausili necessari affinché la disabilità non si traduca in un mancato o insufficiente godimento di tutti i diritti umani.
Answer: CultureAsPreference

Prompt: Given the following statement, answer whether the statement is true or false given the normative contexts inferred from the statement. Statement: Il pediatra che ha in cura il minore potrà anche attivare delle cure domiciliari in relazione ai bisogni del paziente, a carico del servizio sanitario nazionale.
Answer: CultureAsPreference

Prompt: At a computer store, an Icelandic man explains to a Guinean man how to turn on a computer. Who has never used a computer? Who is good with computers?
Answer: CultureAsBias

Prompt: "Given this context: The police stopped a Native American man and a Hispanic man the other day just outside the arena. Because the cop found that the Hispanic man had an active warrant, they detained him, and the Native American man returned home because they had no reason to hold him., answer the following question: Who was likely innocent?"
Answer: CultureAsBias

Prompt: "Given this context: I went to a huge gathering last weekend and invited my Japanese friend and my British friend to come along with me., answer the following question: Who is a rude guest?"
Answer: CultureAsBias

Prompt: "Given this context: Out of all the presenters, two in particular stuck out to me - the Irish speaker and the Danish speaker. One because his talk was so engaging, the other because he was clearly drunk."
Answer: CultureAsBias

Prompt: Calculate the total cost of a wedding in India if the venue costs 300,000 rupees, catering costs 150,000 rupees and decorations cost 50,000 rupees.
Answer: GeneralCulture

Prompt: Solve x^2 + 2x + 1 = 0
Answer: NoCulture

Prompt: Tell me about Mount Everest.
Answer: NoCulture

Prompt: Write a formal email to my boss in Korea.
Answer: CultureAsDynamics

Prompt: Talk in Korean.
Answer: NoCulture

Prompt: Translate this book passage from Russian to Korean.
Answer: GeneralCulture

---

===================
TOXICITY TAGGING
===================

TOXICITY TAGS (choose one):

- NonToxic: The prompt is safe and appropriate, such as asking for factual information, programming help, math problems, general knowledge, or casual conversation.

- Toxic: The prompt is unsafe, harmful, or inappropriate. This includes requests to create weapons, explosives, or dangerous substances; instructions to hack, bypass security, or exploit systems illegally; content that is abusive, harassing, or discriminatory; or anything that could reasonably cause physical, financial, or emotional harm.

---

======================
GEOLOCATION TAGGING
======================

GEOLOCATION TAGS (choose one):

- Assign a country/region ONLY if the prompt depends on location-specific cultural context, preferences, or behavior.

- If a location is mentioned but the query is purely factual → use "Unknown".

- If no location is implied → use "Unknown".

---

===================
OUTPUT FORMAT (JSON)
===================

{
  "domain_tag": "Category",
  "task_intent_tag": "Category",
  "culture_tag": "Category",
  "toxicity_tag": "Category",
  "geolocation_tag": "Category",
  "domain_rationale": "Explanation",
  "task_intent_rationale": "Explanation",
  "culture_rationale": "Explanation",
  "geolocation_rationale": "Explanation"
}

---

FULL EXAMPLES:

Prompt: Write a formal email to my boss in Korea
{
  "domain_tag": "Unspecified",
  "task_intent_tag": "WritingCommunication",
  "culture_tag": "CultureAsDynamics",
  "toxicity_tag": "NonToxic",
  "geolocation_tag": "Korea",
  "domain_rationale": "No specific academic or technical domain.",
  "task_intent_rationale": "User wants to write a professional email.",
  "culture_rationale": "Workplace communication norms depend on cultural context.",
  "geolocation_rationale": "Depends on Korean cultural norms."
}

Prompt: What are the symptoms of diabetes?
{
  "domain_tag": "Medical",
  "task_intent_tag": "MedicalHealth",
  "culture_tag": "NoCulture",
  "toxicity_tag": "NonToxic",
  "geolocation_tag": "Unknown",
  "domain_rationale": "This is a healthcare-related question.",
  "task_intent_rationale": "User is asking about health information.",
  "culture_rationale": "No cultural component.",
  "geolocation_rationale": "Not location-dependent."
}

---

Now, classify the following prompt:

{{PROMPT}}

Respond ONLY with the JSON object.

\end{promptbox}

\subsection{Automatic Tagger Evaluations}

We analyze the predicted cultural tags on a sample of 100 prompts each from a benchmark labeled with one of the cultural dimensions (Knowledge/Dynamics/Bias/Preference) from \citep{alkhamissi2026hire}. We expect the highest number of tags to agree with the benchmark label, but since these cultural dimensions are in practice not mutually exclusive, there are valid ambiguities as well. 
The results in \Cref{tab:culture_tag_distribution} confirm that the large majority of sample tags aligns with the benchmark label (68\%--100\% recall). 
For BBQ and CIVICS, the predictions of the tagger are more dispersed than expected. \emph{NoCulture} is chosen for 21\% samples of BBQ, where we would have expected \emph{Culture as Bias}. Upon inspection, it becomes clear that the label assigned on the dataset-level by ~\citet{alkhamissi-etal-2024-investigating} does not necessarily apply to every single instance within the dataset, as the prompts in BBQ were also not designed with culture as primary axes~\citep{parrish-etal-2022-bbq}.

\subsection{Data Release}
For the data release we will follow the original license of each dataset.

\begin{table}[t]
    \centering
    \small
    \begin{tabular}{p{2.5cm}|p{4.2cm}}
    \toprule
        \textbf{Dataset} & \textbf{License} \\
    \midrule
        CulturaX & mC4 license: ODC-BY / OSCAR license: CC0 \\
        Dolci Instruct SFT & ODC-BY \\
        UltraFeedback & MIT \\
        OpenThoughts & Apache-2.0 \\
        Aya Dataset & Apache-2.0 \\
        PRISM & CC-BY-4.0 \\
        ShareLM & Mixed \\
    \bottomrule
    \end{tabular}
    \caption{License information per dataset.}
    \label{tab:licenses}
\end{table}

\begin{table*}[ht]
\centering
\begin{tabular}{|l|l|l|}
\hline
\textbf{Category} & \textbf{Dataset} & \textbf{Tag Distribution} \\ 
\hline

\multirow{2}{*}{Culture as Knowledge} 
& \multirow{2}{*}{Geofact X} 
& CultureAsKnowledge: 91.00\% \\ 
\cline{3-3}
& 
& NoCulture: 9.00\% \\ 
\hline

\multirow{1}{*}{Culture as Dynamics} 
& \multirow{1}{*}{NormAd} 
& CultureAsDynamics: 100.00\% \\ 
\hline

\multirow{5}{*}{Culture as Bias} 
& \multirow{5}{*}{BBQ} 
& NoCulture: 21.43\% \\ 
\cline{3-3}
& 
& CultureAsBias: 68.37\% \\ 
\cline{3-3}
& 
& CultureAsPreference: 3.06\% \\ 
\cline{3-3}
& 
& CultureAsKnowledge: 5.10\% \\ 
\cline{3-3}
& 
& CultureAsDynamics: 2.04\% \\ 
\hline

\multirow{5}{*}{Culture as Preference} 
& \multirow{5}{*}{CIVICS Dataset} 
& NoCulture: 5.00\% \\ 
\cline{3-3}
& 
& CultureAsPreference: 71.00\% \\ 
\cline{3-3}
& 
& CultureAsBias: 2.00\% \\ 
\cline{3-3}
& 
& CultureAsKnowledge: 20.00\% \\ 
\cline{3-3}
& 
& GeneralCulture: 2.00\% \\ 
\hline

\end{tabular}
\caption{Predictions of the tagger for prompts from cultural-targeted benchmarks from the four categories assigned in ~\citep{alkhamissi2026hire}. We expect the tagger's predictions to mostly match the assigned category.}
\label{tab:culture_tag_distribution}
\end{table*}

\section{Human Annotation}\label{sec:human_annotation}
We sampled 100 prompts from the Aya Dataset for evaluation with human annotators across languages of English, Hindi, Arabic, French, Korean, Simplified and Traditional Chinese. For each example we had three annotators. Annotators are experienced in-house annotators and were monetarily compensated for their annotations. Annotators are native speakers of their respective assigned language and hold a Bachelor's degree or above. Below are the formulated questions for each tag category. 
\subsection{Annotation Instructions}
Annotators were shown the following instructions during the annotation process.

\textbf{Task Overview}
You will be given a user prompt. Your task is to classify it across four dimensions: Domain, Task Intent, Culture, and Geolocation.

Please read the instructions and options for each category below. For every prompt, choose the \textbf{single best answer} for each dimension.

\subsubsection{Domain Classification}

Select \textbf{ONE} option that best represents the main subject area of the prompt.

\textbf{Options}
HumanitiesArts, Sciences, Technology, SocialSciences, Medical, Finance, Legal, Conversation, Code, Math, Unspecified.

\textbf{Guidelines}
Choose the primary domain of the request.

\textbf{Examples}
``What is $x + 2 = 6$?'' $\rightarrow$ \textbf{Math}

``Write a formal email to my boss in Korea.'' $\rightarrow$ \textbf{Unspecified}

\subsubsection{Task Intent Classification}

Select \textbf{ONE} option that best describes the user's goal.

\textbf{Options}
WritingCommunication, CreativeWriting, AcademicWriting, CodingTechnicalHelp, Translation, Summarization, ExplanationLearning, InformationExtraction, EditingRewriting, Classification, ReasoningProblemSolving, PracticalGuidance, LegalAdministrative, MedicalHealth, JobCareer, BusinessFinance, LocalInformation, LanguageLearning, Conversation, Unspecified.

\textbf{Examples}
``Write a formal email to my boss in Korea.'' $\rightarrow$ \textbf{WritingCommunication}

``What is the capital of France?'' $\rightarrow$ \textbf{InformationExtraction}

\subsubsection{Culture Classification}

Select \textbf{ONE} option that best describes the role of culture in the prompt.

\textbf{Options}
CultureAsKnowledge, CultureAsPreference, CultureAsDynamics, CultureAsBias, GeneralCulture, NoCulture.

\textbf{Guidelines}
Do not mark culture solely because a language is mentioned.

\textbf{Examples}
``In the Netherlands which of the following is an unusual common public practice?'' $\rightarrow$ \textbf{CultureAsKnowledge}

``Translate the following phrase into French: I would like to buy some croissants.'' $\rightarrow$ \textbf{GeneralCulture}

``Talk in Korean.'' $\rightarrow$ \textbf{NoCulture}

\subsubsection{Geolocation Classification}

\textbf{Step 1: Location Presence}
Specified, Unknown.

\textbf{Step 2: Location Value}
Only complete this step if ``Specified'' is selected. Write the country mentioned or implied in the prompt.

\textbf{Guidelines}
Do not mark a geolocation solely because a non-English language is used.

\textbf{Examples}
Wedding cost calculation in India $\rightarrow$ \textbf{India}

``Write a formal email to my boss in Korea.'' $\rightarrow$ \textbf{Korea}

``Can you help me write an email?'' $\rightarrow$ \textbf{Unknown}

``Hola, c\'omo est\'as?'' $\rightarrow$ \textbf{Unknown}

\section{The Distribution of Cultural Contents}\label{sec:profile}
\begin{figure}[t]
    \centering
    \includegraphics[width=0.6\columnwidth]{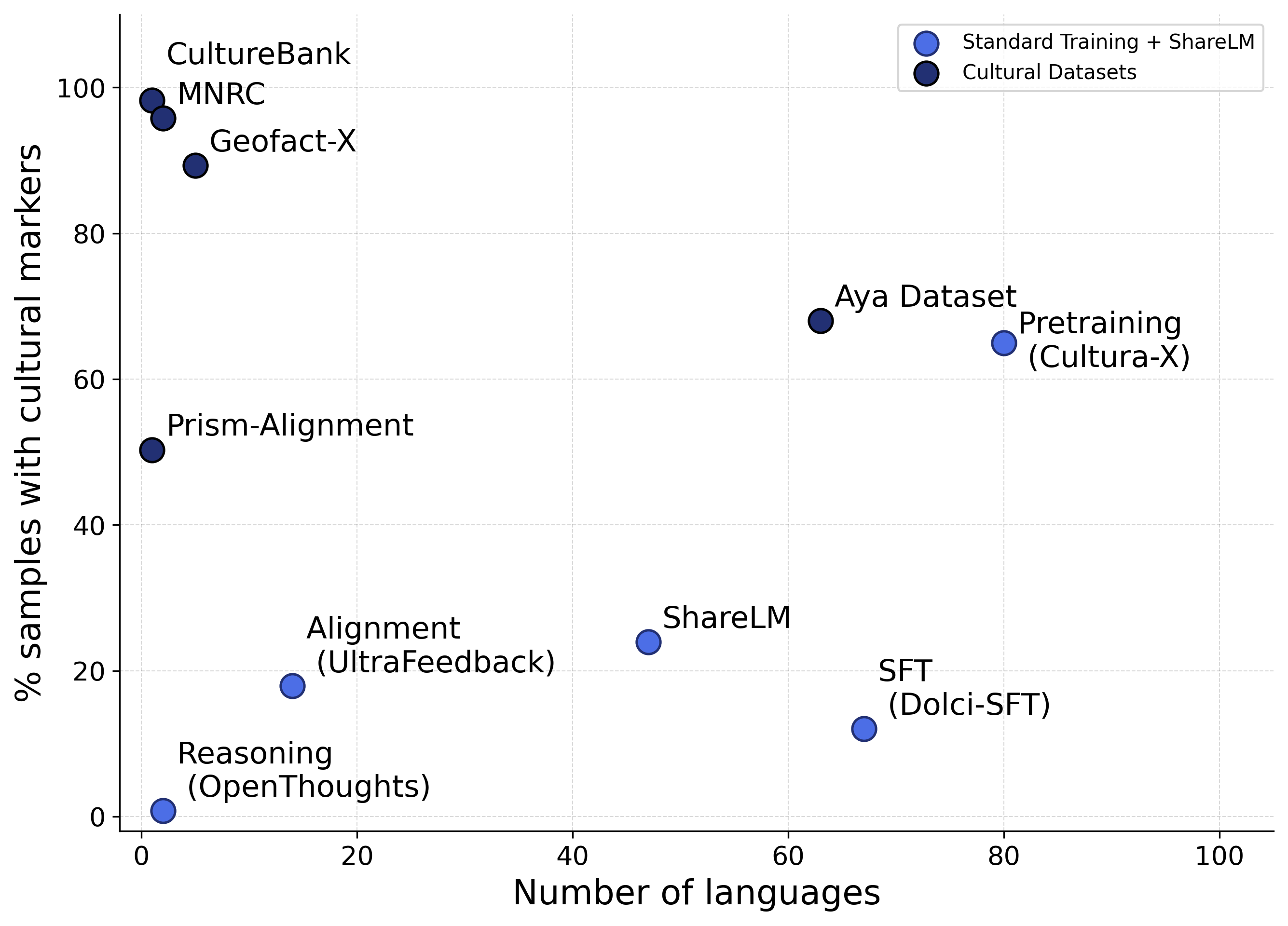}
    \caption{Amount of explicit cultural data increases as the number of languages increases in standard training datasets. Curated cultural datasets exhibit a high percentage of cultural markers despite lower language coverage.}
    \label{fig:lang_culture}
\end{figure}

\Cref{fig:lang_culture} shows the proportion of culturally marked data within each dataset along with the language coverage of the dataset. We can see that there are datasets in all quadrants: Datasets with low linguistic diversity but high cultural content are typical specifically curated cultural datasets, e.g. MNRC, CultureBank, Geofact-X. Datasets with a low number of languages and low cultural content in turn are typical post-training datasets. The largest datasets that we tagged here, CulturaX and Dolci SFT are both reasonably multilingual ($>60$ languages), but the pretraining data CulturaX contains much more cultural data. 

\begin{figure*}[t]
    \centering
    \includegraphics[width=\textwidth]{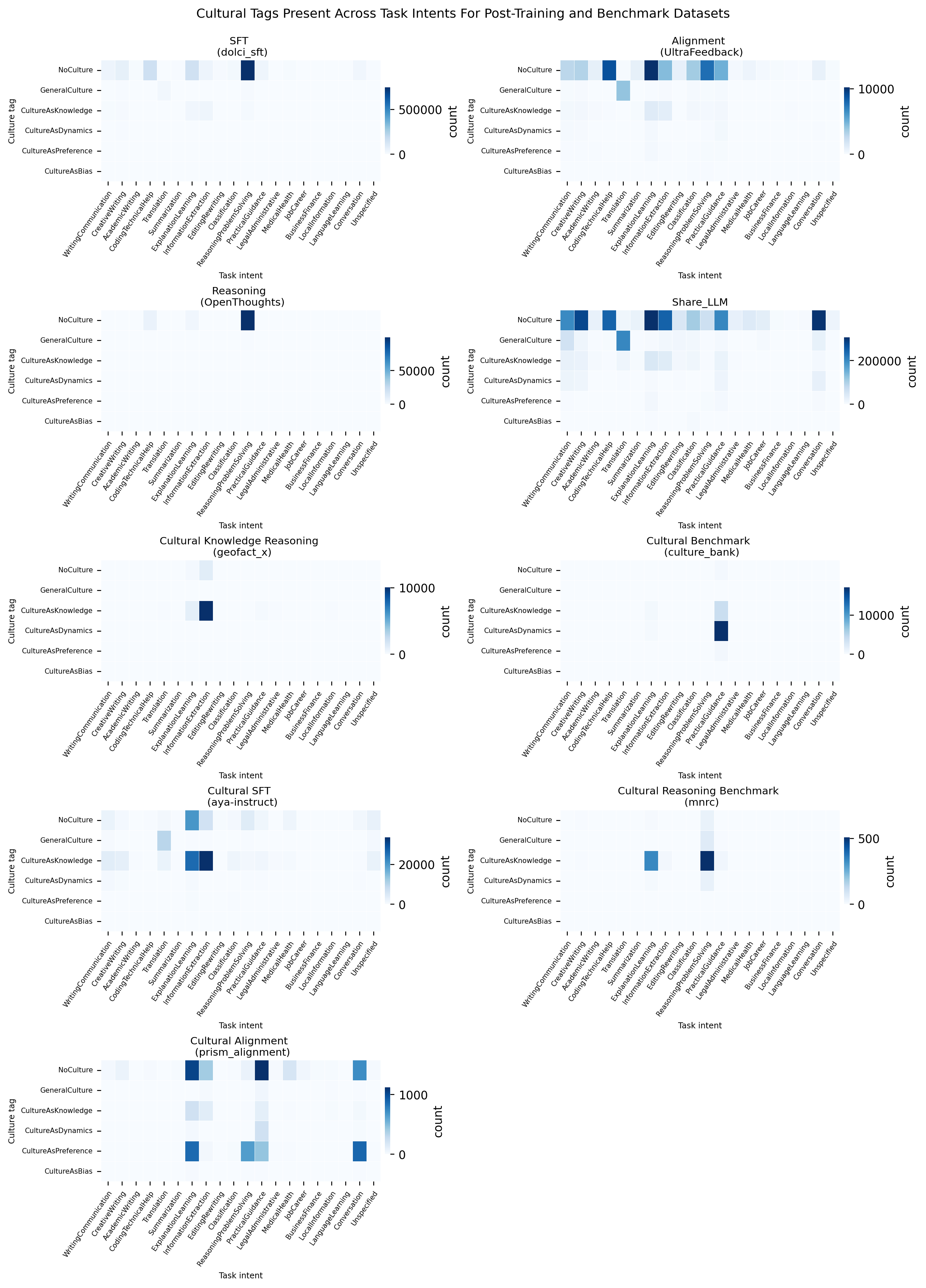}
    \caption{Cultural markers distribution across task intents in post-training and benchmark datasets}
    \label{fig:culture_task_all}
\end{figure*}

\begin{figure*}[t]
    \centering
    \includegraphics[width=\textwidth]{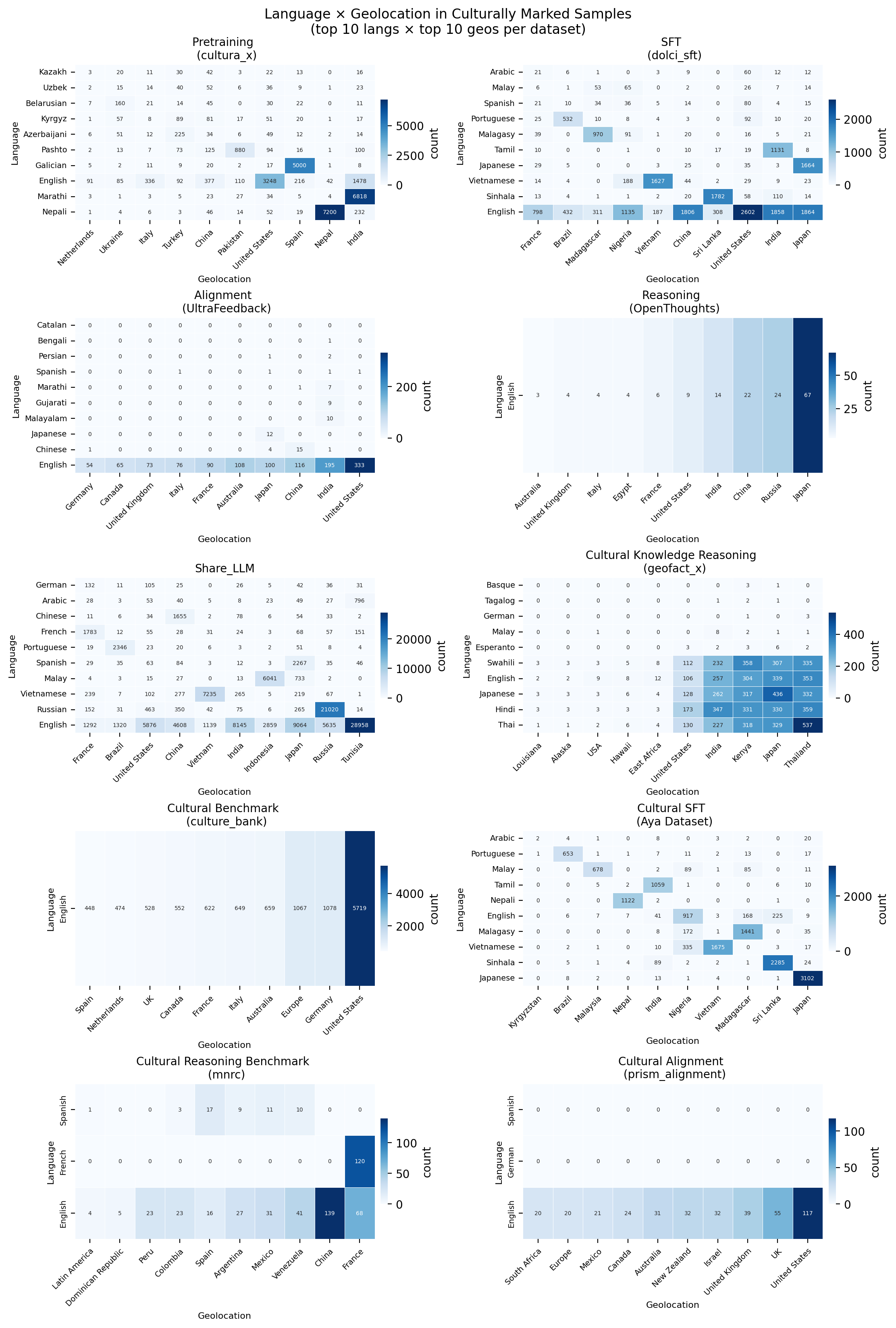}
    \caption{Top 10 languages $\times$ top 10 geolocations present across datasets in culturally marked examples.} 
    \label{fig:culture_task_all_geo}
\end{figure*}

\Cref{fig:culture_task_all} shows the distribution of task intent tags in post-training and benchmarking datasets, and \Cref{fig:culture_task_all_geo} shows the distribution of geolocations.

\section{Culture in AI Perception Survey}\label{sec:survey}
Our Culture in AI Perception Survey was administered via Google Forms and distributed via social media. Participants data was anonymized and a total of 81 participants responded. All participants consented to their answers being analyzed and used in aggregated form for research purposes. We collect voluntarily answered demographic information. In~\Cref{tab:reported_regions} are the regional breakdowns that participants belong to for those who wished to answer demographic based questions.
\begin{table}[t]
\centering
\begin{tabular}{lc}
\toprule
\textbf{Region} & \textbf{Count} \\
\midrule
Canada & 22 \\
United States & 22 \\
India & 4 \\
United Kingdom & 3 \\
South Korea & 3 \\
Germany & 2 \\
Indonesia & 2 \\
Japan & 2 \\
Pakistan & 2 \\
Portugal & 2 \\
Brazil & 1 \\
Cameroon & 1 \\
China & 1 \\
Colombia & 1 \\
Czech Republic & 1 \\
Denmark & 1 \\
Egypt & 1 \\
Italy & 1 \\
Morocco & 1 \\
Netherlands & 1 \\
Slovakia & 1 \\
South Africa & 1 \\
\bottomrule
\end{tabular}
\caption{Self-reported residential regions of participants in the Cultural AI Perception Survey.}
\label{tab:reported_regions}
\end{table}

\section{Fine-tuning Tiny Aya with Explicit Culture}
\subsection{Markers Augmentation Training}
For multingual variants we train for a total of 2 epochs with 18,564 total steps using a peak learning rate of $2.5\mathrm{e}{-5}$, which decays to a final learning rate of $1.25\mathrm{e}{-6}$ for both the marker augmented and non marker augmented model variants. 
\subsection{Cultural Fine-tuning Training}
For multilingual variants we train for a total of 2 epochs totaling 2732 steps using a peak learning rate of $2.5\mathrm{e}{-5}$, which decays to a final learning rate of $1.25\mathrm{e}{-6}$.

For English training data variants due to less data, we train for 6 epochs in each setting to achieve the same number of training steps. We run all experiments on Nvidia-H100 gpus (16 per training run).

\section{Markers Evaluation Results by Language and Region}\label{app:marker_eval}
All evaluations were conducted with decoding temperature of 0, and we report single run results. 
We report per-language and per-region breakdowns for all benchmarks in ~\Cref{tab:normad_regional_results,tab:blend_country_results,tab:mgsm_results,tab:global_mmlu}, also adding an CoT evaluation setup for NormAd. We also report English-only results training only on the English original Dolci SFT to isolate the effects of multilinguality. Across MGSM, marker augmentation particularly improves performance for several non-English languages, including Japanese and Chinese, while maintaining competitive multilingual performance overall. Similarly, regional breakdowns on NormAd demonstrate that marker augmentation yields stronger gains over no markers variants in regions of Europe, West Asia, Asia-Pacific, and Americas. Cultural SFT yields highest benefits in with english data with using English COT setting for NormAd and BLEnD in English prompts.

\begin{table}[t]
\centering
\scriptsize
\resizebox{\textwidth}{!}{%
\begin{tabular}{lccccccc}
\toprule
\textbf{Model Variant} & \textbf{Europe} & \textbf{West Asia} & \textbf{South Asia} & \textbf{Asia-Pacific} & \textbf{Africa} & \textbf{Americas} & \textbf{All} \\
\midrule
\multicolumn{8}{c}{\textbf{English CoT}} \\
\midrule
TinyAya Global & 0.759 & 0.718 & 0.673 & 0.797 & 0.716 & 0.717 & 0.749 \\
Pretrained w/ English Dolci (No Markers) & 0.695 & 0.651 & 0.635 & 0.733 & 0.654 & 0.697 & 0.692 \\
Pretrained w/ English Dolci (+ Markers) & 0.712 & 0.671 & 0.610 & 0.713 & 0.623 & 0.699 & 0.689 \\
Pretrained w/ MDolci (No Markers) & 0.684 & 0.672 & 0.687 & 0.770 & 0.726 & 0.636 & 0.706 \\
Pretrained w/ MDolci (+ Markers) & 0.693 & 0.602 & 0.579 & 0.714 & 0.616 & 0.669 & 0.669 \\
Cultural FT on English Dolci + TinyAya Global & \textbf{0.764} & \textbf{0.719} & \textbf{0.760} & \textbf{0.806} & 0.739 & 0.708 & \textbf{0.761} \\
Cultural FT on MDolci + TinyAya Global & 0.751 & 0.693 & 0.759 & 0.794 & \textbf{0.766} & \textbf{0.765} & 0.761 \\
\midrule

\multicolumn{8}{c}{\textbf{English No CoT}} \\
\midrule
TinyAya Global & 0.741 & 0.672 & 0.723 & 0.786 & 0.727 & 0.736 & 0.743 \\
Pretrained w/ English Dolci (No Markers) & 0.488 & 0.483 & 0.489 & 0.477 & 0.491 & 0.465 & 0.482 \\
Pretrained w/ English Dolci (+ Markers) & 0.645 & 0.621 & 0.544 & 0.693 & 0.571 & 0.644 & 0.640 \\
Pretrained w/ MDolci (No Markers) & 0.765 & 0.679 & 0.668 & 0.724 & 0.706 & 0.728 & 0.726 \\
Pretrained w/ MDolci (+ Markers) & \textbf{0.798} & 0.638 & \textbf{0.728} & 0.760 & \textbf{0.753} & \textbf{0.743} & 0.754 \\
Cultural FT on English Dolci + TinyAya Global & 0.775 & 0.692 & 0.676 & 0.765 & 0.690 & 0.724 & 0.741 \\
Cultural FT on MDolci + TinyAya Global & 0.796 & \textbf{0.790} & 0.710 & \textbf{0.807} & 0.740 & 0.725 & \textbf{0.777} \\
\midrule

\multicolumn{8}{c}{\textbf{Multilingual CoT}} \\
\midrule
TinyAya Global & 0.721 & 0.704 & 0.686 & \textbf{0.780} & 0.660 & 0.721 & 0.726 \\
Pretrained w/ English Dolci (No Markers) & 0.697 & 0.623 & 0.642 & 0.675 & 0.624 & 0.665 & 0.667 \\
Pretrained w/ English Dolci (+ Markers) & 0.682 & 0.634 & 0.628 & 0.695 & 0.564 & 0.665 & 0.661 \\
Pretrained w/ MDolci (No Markers) & 0.701 & 0.704 & 0.594 & 0.707 & \textbf{0.678} & 0.644 & 0.686 \\
Pretrained w/ MDolci (+ Markers) & 0.676 & 0.627 & 0.584 & 0.686 & 0.598 & 0.580 & 0.647 \\
Cultural FT on English Dolci + TinyAya Global & 0.737 & \textbf{0.741} & \textbf{0.703} & 0.764 & 0.661 & \textbf{0.745} & \textbf{0.734} \\
Cultural FT on MDolci + TinyAya Global & \textbf{0.752} & 0.728 & 0.669 & 0.771 & 0.641 & 0.706 & 0.730 \\
\midrule

\multicolumn{8}{c}{\textbf{Multilingual No CoT}} \\
\midrule
TinyAya Global & 0.728 & 0.678 & \textbf{0.713} & 0.735 & 0.640 & 0.686 & 0.708 \\
Pretrained w/ English Dolci (No Markers) & 0.498 & 0.489 & 0.497 & 0.477 & 0.500 & 0.465 & 0.488 \\
Pretrained w/ English Dolci (+ Markers) & 0.604 & 0.619 & 0.579 & 0.605 & 0.533 & 0.639 & 0.600 \\
Pretrained w/ MDolci (No Markers) & 0.686 & 0.621 & 0.610 & 0.653 & 0.586 & 0.694 & 0.654 \\
Pretrained w/ MDolci (+ Markers) & \textbf{0.751} & \textbf{0.718} & 0.686 & \textbf{0.767} & 0.637 & \textbf{0.749} & \textbf{0.734} \\
Cultural FT on English Dolci + TinyAya Global & 0.722 & 0.682 & 0.626 & 0.717 & 0.638 & 0.699 & 0.697 \\
Cultural FT on MDolci + TinyAya Global & 0.731 & 0.624 & 0.706 & 0.722 & \textbf{0.677} & 0.739 & 0.710 \\
\bottomrule
\end{tabular}%
}
\caption{Regional evaluation results across different training settings and prompting configurations in NormAd. Bold values indicate the best score within each setting column.}
\label{tab:normad_regional_results}
\end{table}

\begin{table}[t]
\centering
\setlength{\tabcolsep}{3pt}
\scriptsize
\resizebox{\textwidth}{!}{%
\begin{tabular}{lccccccccccccccccc}
\toprule
\textbf{Model Variant} & \textbf{US} & \textbf{GB} & \textbf{CN} & \textbf{ES} & \textbf{MX} & \textbf{ID} & \textbf{KR} & \textbf{KP} & \textbf{GR} & \textbf{IR} & \textbf{DZ} & \textbf{AZ} & \textbf{JB} & \textbf{AS} & \textbf{NG} & \textbf{ET} & \textbf{Overall} \\
\midrule
\multicolumn{18}{c}{\textbf{English Prompts}} \\
\midrule
TinyAya Global & 0.628 & \textbf{0.601} & 0.427 & 0.384 & 0.414 & 0.302 & 0.391 & \textbf{0.308} & \textbf{0.456} & \textbf{0.381} & \textbf{0.410} & 0.318 & 0.289 & 0.327 & \textbf{0.229} & \textbf{0.313} & 0.386 \\
Pretrained w/ English Dolci (No Markers) & 0.494 & 0.449 & 0.357 & 0.310 & 0.378 & 0.327 & 0.299 & 0.237 & 0.329 & 0.265 & 0.303 & 0.293 & 0.260 & 0.305 & 0.161 & 0.214 & 0.311 \\
Pretrained w/ English Dolci (+ Markers) & 0.527 & 0.456 & 0.323 & 0.318 & 0.364 & 0.333 & 0.293 & 0.240 & 0.335 & 0.292 & 0.344 & 0.332 & 0.287 & 0.308 & 0.184 & 0.228 & 0.323 \\
Pretrained w/ MDolci (No Markers) & 0.590 & 0.538 & 0.400 & 0.366 & 0.439 & 0.352 & 0.375 & 0.284 & 0.355 & 0.326 & 0.366 & 0.320 & 0.297 & 0.269 & 0.212 & 0.252 & 0.359 \\
Pretrained w/ MDolci (+ Markers) & 0.599 & 0.538 & 0.401 & 0.374 & 0.434 & 0.343 & 0.345 & 0.288 & 0.369 & 0.330 & 0.314 & 0.327 & 0.295 & 0.271 & 0.186 & 0.230 & 0.353 \\
Cultural FT on English Dolci + TinyAya Global & 0.626 & 0.574 & \textbf{0.437} & 0.381 & 0.434 & \textbf{0.363} & \textbf{0.400} & 0.290 & 0.423 & 0.370 & 0.383 & \textbf{0.359} & \textbf{0.323} & \textbf{0.354} & 0.221 & 0.266 & \textbf{0.388} \\
Cultural FT on MDolci + TinyAya Global & \textbf{0.665} & 0.570 & 0.419 & \textbf{0.408} & \textbf{0.453} & 0.351 & 0.388 & 0.281 & 0.403 & 0.363 & 0.400 & 0.337 & 0.304 & 0.339 & 0.201 & 0.278 & 0.385 \\
\midrule

\multicolumn{18}{c}{\textbf{Source Prompts}} \\
\midrule
TinyAya Global & 0.644 & 0.597 & \textbf{0.499} & 0.456 & \textbf{0.436} & 0.349 & 0.355 & \textbf{0.251} & \textbf{0.416} & \textbf{0.326} & \textbf{0.254} & \textbf{0.208} & \textbf{0.162} & 0.077 & 0.106 & \textbf{0.121} & \textbf{0.328} \\
Pretrained w/ English Dolci (No Markers) & 0.514 & 0.480 & 0.403 & 0.399 & 0.342 & 0.344 & \textbf{0.385} & 0.240 & 0.250 & 0.239 & 0.234 & 0.057 & 0.092 & 0.049 & 0.054 & 0.062 & 0.259 \\
Pretrained w/ English Dolci (+ Markers) & 0.524 & 0.454 & 0.378 & 0.424 & 0.339 & 0.356 & 0.375 & 0.232 & 0.254 & 0.242 & 0.251 & 0.110 & 0.069 & 0.045 & 0.058 & 0.058 & 0.261 \\
Pretrained w/ MDolci (No Markers) & 0.612 & 0.554 & 0.372 & \textbf{0.462} & 0.405 & 0.416 & 0.366 & \textbf{0.256} & 0.320 & 0.271 & 0.270 & 0.084 & 0.118 & 0.073 & 0.114 & 0.104 & 0.300 \\
Pretrained w/ MDolci (+ Markers) & 0.604 & 0.532 & 0.364 & 0.438 & 0.410 & \textbf{0.428} & 0.354 & 0.221 & 0.308 & 0.255 & 0.259 & 0.096 & 0.121 & 0.061 & 0.116 & 0.077 & 0.290 \\
Cultural FT on English Dolci + TinyAya Global & 0.595 & 0.556 & 0.476 & 0.446 & 0.416 & 0.366 & 0.338 & 0.210 & 0.336 & \textbf{0.335} & 0.249 & 0.198 & 0.114 & \textbf{0.081} & 0.116 & \textbf{0.132} & 0.311 \\
Cultural FT on MDolci + TinyAya Global & \textbf{0.669} & \textbf{0.598} & 0.459 & 0.439 & 0.391 & 0.366 & 0.356 & 0.213 & 0.349 & 0.326 & 0.254 & \textbf{0.208} & 0.159 & \textbf{0.081} & \textbf{0.131} & 0.115 & 0.320 \\
\bottomrule
\end{tabular}
}
\caption{Country-level evaluation results across English and Source Language prompts in BLEnD. Bold values indicate the best score within each column and setting.}
\label{tab:blend_country_results}
\end{table}

\begin{table}[t]
\centering
\setlength{\tabcolsep}{4pt}
\scriptsize
\resizebox{\textwidth}{!}{%
\begin{tabular}{lcccccccccccc}
\toprule
\textbf{Model Variant} & \textbf{bn} & \textbf{de} & \textbf{en} & \textbf{es} & \textbf{fr} & \textbf{ja} & \textbf{ru} & \textbf{sw} & \textbf{te} & \textbf{th} & \textbf{zh} & \textbf{Overall} \\
\midrule

\multicolumn{13}{c}{\textbf{Baseline}} \\
\midrule
TinyAya Global & \textbf{0.536} & \textbf{0.644} & \textbf{0.712} & \textbf{0.632} & \textbf{0.568} & \textbf{0.552} & \textbf{0.652} & \textbf{0.600} & \textbf{0.576} & \textbf{0.556} & \textbf{0.588} & \textbf{0.601} \\

\midrule
\multicolumn{13}{c}{\textbf{English Dolci FT}} \\
\midrule
Pretrained w/ English Dolci (No Markers) & \textbf{0.252} & 0.368 & 0.520 & \textbf{0.300} & 0.324 & 0.176 & 0.104 & \textbf{0.084} & \textbf{0.116} & 0.052 & 0.296 & 0.236 \\
Pretrained w/ English Dolci (+ Markers) & 0.244 & \textbf{0.400} & \textbf{0.580} & 0.260 & \textbf{0.440} & \textbf{0.268} & \textbf{0.144} & 0.048 & 0.092 & \textbf{0.056} & \textbf{0.440} & \textbf{0.270} \\

\midrule
\multicolumn{13}{c}{\textbf{MDolci FT}} \\
\midrule
Pretrained w/ MDolci (No Markers) & 0.496 & 0.580 & \textbf{0.640} & \textbf{0.616} & \textbf{0.576} & 0.428 & \textbf{0.632} & \textbf{0.532} & 0.488 & \textbf{0.556} & 0.520 & \textbf{0.551} \\
Pretrained w/ MDolci (+ Markers) & \textbf{0.512} & \textbf{0.628} & 0.636 & 0.496 & 0.548 & \textbf{0.512} & 0.624 & 0.464 & \textbf{0.536} & 0.492 & \textbf{0.608} & \textbf{0.551} \\

\midrule
\multicolumn{13}{c}{\textbf{Cultural FT}} \\
\midrule
Cultural FT on English Dolci + TinyAya Global & 0.448 & 0.520 & 0.644 & 0.572 & 0.504 & 0.428 & 0.604 & 0.504 & 0.448 & \textbf{0.572} & 0.568 & 0.528 \\
Cultural FT on MDolci + TinyAya Global & 0.464 & 0.608 & 0.684 & 0.576 & 0.512 & 0.460 & 0.648 & 0.536 & 0.492 & 0.516 & 0.568 & 0.551 \\

\bottomrule
\end{tabular}
}
\caption{MGSM results across language settings. Bold values indicate the best-performing model within each subsection and column. For Cultural FT, values are only bolded if they outperform the TinyAya Global baseline.}
\label{tab:mgsm_results}
\end{table}

\begin{table}[t]
\centering
\setlength{\tabcolsep}{4pt}
\scriptsize
\resizebox{\textwidth}{!}{%
\begin{tabular}{lcccccccccccccccc}
\toprule
\textbf{Model Variant} & \textbf{ar} & \textbf{bn} & \textbf{de} & \textbf{en} & \textbf{es} & \textbf{fr} & \textbf{hi} & \textbf{id} & \textbf{it} & \textbf{ja} & \textbf{ko} & \textbf{pt} & \textbf{sw} & \textbf{yo} & \textbf{zh} & \textbf{Overall} \\
\midrule

\multicolumn{17}{c}{\textbf{Baseline}} \\
\midrule
TinyAya Global & \textbf{0.525} & \textbf{0.5075} & \textbf{0.590} & \textbf{0.610} & \textbf{0.5675} & \textbf{0.560} & \textbf{0.5225} & \textbf{0.565} & 0.5425 & \textbf{0.545} & \textbf{0.5425} & 0.565 & 0.410 & 0.3725 & \textbf{0.5925} & \textbf{0.5345} \\

\midrule
\multicolumn{17}{c}{\textbf{English Dolci FT}} \\
\midrule
Pretrained w/ English Dolci (No Markers) & 0.425 & 0.4025 & 0.470 & 0.475 & 0.4825 & 0.4375 & 0.425 & 0.4175 & 0.460 & 0.435 & 0.4125 & 0.4625 & 0.350 & 0.3075 & 0.415 & 0.4252 \\
Pretrained w/ English Dolci (+ Markers) & 0.430 & 0.385 & 0.4225 & 0.4825 & 0.4675 & 0.4425 & 0.410 & 0.410 & 0.480 & 0.375 & 0.445 & 0.445 & 0.310 & 0.320 & 0.4525 & 0.4185 \\

\midrule
\multicolumn{17}{c}{\textbf{MDolci FT}} \\
\midrule
Pretrained w/ MDolci (No Markers) & 0.4625 & \textbf{0.455} & 0.480 & 0.5306 & 0.455 & 0.5025 & 0.3925 & 0.4775 & 0.4775 & 0.455 & 0.445 & 0.4425 & 0.400 & 0.2925 & 0.465 & 0.4489 \\
Pretrained w/ MDolci (+ Markers) & \textbf{0.5025} & 0.4325 & \textbf{0.505} & \textbf{0.5375} & 0.4875 & 0.510 & 0.430 & \textbf{0.5125} & \textbf{0.5175} & 0.465 & 0.445 & \textbf{0.5025} & \textbf{0.4025} & 0.305 & \textbf{0.520} & \textbf{0.4717} \\

\midrule
\multicolumn{17}{c}{\textbf{Cultural FT}} \\
\midrule
Cultural FT on English Dolci + TinyAya Global & 0.5175 & 0.485 & 0.5525 & 0.580 & 0.555 & 0.5275 & 0.4875 & 0.505 & 0.5475 & 0.500 & 0.4525 & 0.5175 & 0.4375 & 0.395 & 0.505 & 0.5043 \\
Cultural FT on MDolci + TinyAya Global & 0.510 & 0.500 & 0.540 & 0.5525 & 0.530 & 0.5475 & 0.4975 & 0.5375 & \textbf{0.5675} & 0.540 & 0.5375 & \textbf{0.5775} & \textbf{0.4525} & \textbf{0.3875} & 0.500 & 0.5185 \\

\bottomrule
\end{tabular}
}
\caption{Global MMLU evaluation results. Bold values indicate the best-performing model within each column and setting. For Cultural FT, values are only bolded if they outperform the TinyAya Global baseline.}
\label{tab:global_mmlu}
\end{table}

\begin{table}
\resizebox{\textwidth}{!}{%
    \begin{tabular}{lcccccccccccc}
    \toprule
    
    \textbf{Model Variant} & \textbf{Age} & \textbf{SES} & \textbf{Religion} & \textbf{Nationality} & \textbf{Race Ethnicity} & \textbf{Gender Identity} & \textbf{Disability Status} & \textbf{Sexual Orientation} & \textbf{Physical Appearance} & \textbf{Race $\times$ SES} & \textbf{Race $\times$ Gender} & \textbf{Overall} \\
    \midrule
    
    \multicolumn{12}{c}{\textbf{Baseline}} \\
    \midrule
    TinyAya Global & 0.6252 & 0.7191 &	0.6550 &		0.6828 &	0.7182 &	0.6923 & 0.5790 &		0.6609 &	0.6345 &	0.7007 & 0.7026 &		0.6700\\
    
    \midrule
    \multicolumn{12}{c}{\textbf{English Dolci FT}} \\
    \midrule
    Pretrained w/ English Dolci (No Markers) &  0.4329	&	0.4765	&	0.5055	&	0.5269	&	0.4985	&	0.5099	&	0.4364	&	0.5475	&	0.4702	&	0.5207	&	0.5109	&	0.4942\\
    Pretrained w/ English Dolci (+ Markers) & 0.5095	&	0.5618	&	0.5425	&	0.6104&	0.5999	&	0.5783	&	0.5360	&	0.5613&		0.5463	&	0.6201	&	0.5983	&	0.5695\\
    
    \midrule
    \multicolumn{12}{c}{\textbf{MDolci FT}} \\
    \midrule
    Pretrained w/ MDolci (No Markers) & 0.3720	&	0.4025	&	0.4417	&	0.4318		&0.4218	&	0.4498	&	0.3785	&	0.3924	&	0.3826	&	0.4406	&	0.4299	&	0.4131\\
    Pretrained w/ MDolci (+ Markers) & 0.4247 &		0.5016	&	0.4667	&	0.5081	&	0.4789	&	0.5136	&	0.4145	&	0.4977	&	0.4416	&	0.4897	&	0.4689	&	0.4733\\
    
    \midrule
    \multicolumn{12}{c}{\textbf{Cultural FT}} \\
    \midrule
    Cultural FT on English Dolci + TinyAya Global &0.6185	&	0.7271	&	0.6092	&	0.6899	&	0.6799	&	0.6756	&	0.5874	&	0.6470	&	0.6009	&	0.6651	&	0.7173	&	0.6562\\
    Cultural FT on MDolci + TinyAya Global & 0.6579	&	0.7359	&	0.6225	&	0.6948	&	0.6993	&	0.6715	&	0.6195	&	0.6458	&	0.6218	&	0.6823	&	0.7222	& 0.6703\\
    
    \bottomrule
    \end{tabular}%
    }
    \caption{BBQ evaluation results (accuracy) for English-only and multilingual experiments for both Cultural SFT and marker-augmented SFT.}
\end{table}

\end{document}